\documentclass[journal]{IEEEtran}
\usepackage{multirow}

\usepackage[pdftex]{graphicx}
\usepackage[cmex10]{amsmath}
\usepackage{algorithmic}
\usepackage{array}
\usepackage{url}
\usepackage{color}

\newcommand{\etal}{\mbox{\emph{et al.\ }}}
\newcommand{\ie}{\mbox{\emph{i. e.\ }}}

\begin{document}
\title{An equalized global graph model-based approach for multi-camera object tracking}

\author{Weihua~Chen*,
        Lijun~Cao*,
        Xiaotang~Chen,~\IEEEmembership{Member,~IEEE},
        and~Kaiqi~Huang$^\dag$,~\IEEEmembership{Senior Member,~IEEE}
\thanks{
*Weihua~Chen and Lijun~Cao contributed equally to this work.

$^\dag$Kaiqi Huang is the correspondence author.

The authors are with the Center for Research on Intelligent Perception and Computing,
National Laboratory of Pattern Recognition,
Institute of Automation, Chinese Academy of Sciences,
No.95 ZhongGuanCun East St, HaiDian District, Beijing, P.R.China, 100190. E-mail: \{weihua.chen,lijun.cao,xtchen,kqhuang\}@nlpr.ia.ac.cn.}
}



\maketitle

\begin{abstract}
Non-overlapping multi-camera visual object tracking
typically consists of two steps: single camera object
tracking and inter-camera object tracking. Most of tracking methods focus on single camera object tracking, which happens in the same scene, while for real surveillance scenes, inter-camera object tracking is needed and single camera tracking methods can not work effectively.
In this paper, we try to improve the overall multi-camera object tracking performance by a global graph model with an improved similarity metric. Our method treats the similarities of single camera tracking and inter-camera tracking differently and obtains the optimization in a global graph model. The results show that our method can work better even in the condition of poor single camera object tracking.
\end{abstract}


%
\IEEEpeerreviewmaketitle

\section{Introduction}
\label{sec:introduction}
\IEEEPARstart{T}{racking} objects of interest is an important and challenging problem in
 intelligent visual surveillance systems~\cite{Vezzani/acmcs2013}.
Since the visual surveillance systems provide huge amount of video streams, it is desirable that objects of interest can be automatically tracked by algorithms instead of human.
Visual object tracking~\cite{Smeulders/pami2014} is a long-standing problem in computer vision, and there are a great
amount of efforts made in visual object tracking within single cameras~\cite{Pang/iccv2013,Wu/cvpr2013,huang2008real}.
In intelligent visual surveillance systems~\cite{Huang/prl2010,Venetianer/cviu2010},
due to the finite camera field of view, it is difficult to observe the complete trajectory
of objects of interest in wide areas with only one camera.
Hence, it is desired to enable the intelligent visual surveillance system
to track the objects of interest within multiple cameras~\cite{Wang/prl2012}.
In addition, for practical considerations, the intelligent visual surveillance
system usually holds the cameras installed with no overlapping areas.
Thus, the intelligent visual surveillance system should be able to track objects of
interest across multiple non-overlapping cameras.
In this paper, we focus on addressing the problem of tracking objects of interest
across multiple non-overlapping cameras.

\begin{figure}[!t]
  \centering
  \centerline{\includegraphics[width=1.0\linewidth]{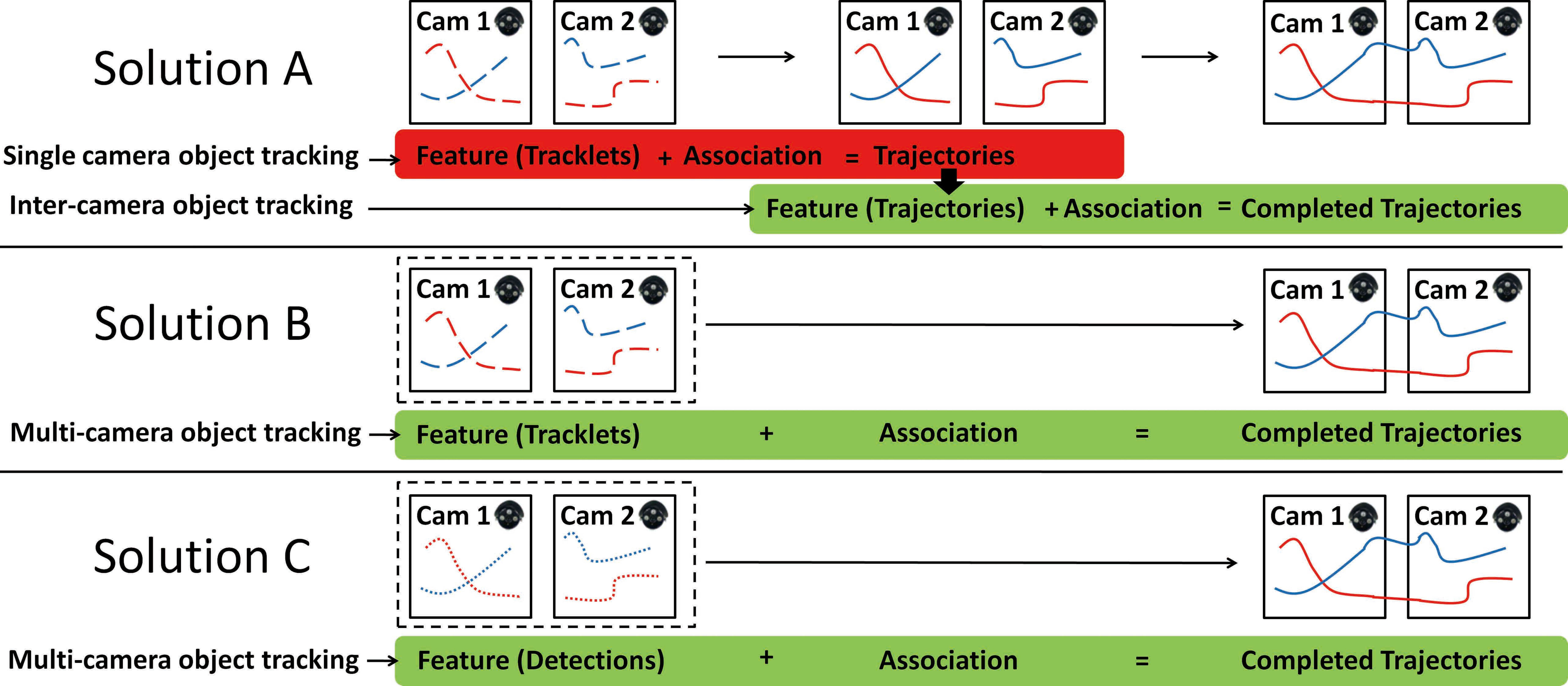}}
%
\caption{\textbf{Illustration of three types of multi-camera visual object tracking solution.}}
\label{fig:solution}
\end{figure}

\begin{figure}[t]
\begin{minipage}[b]{.48\linewidth}
  \centering
  \centerline{\includegraphics[width=1.0\linewidth]{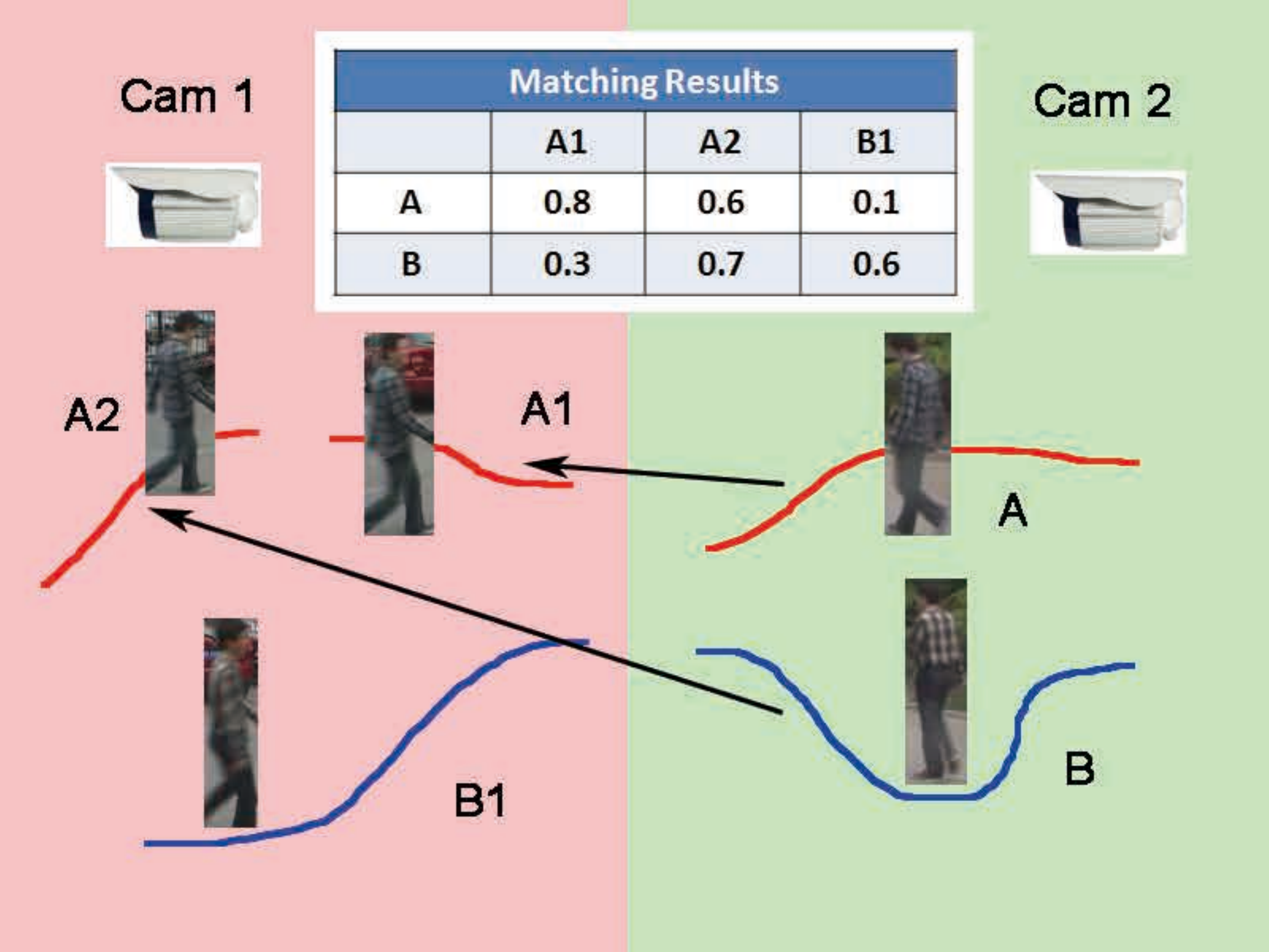}}
  \centerline{(a) Wrong matching}\medskip
\end{minipage}
\hfill
\begin{minipage}[b]{0.48\linewidth}
  \centering
  \centerline{\includegraphics[width=1.0\linewidth]{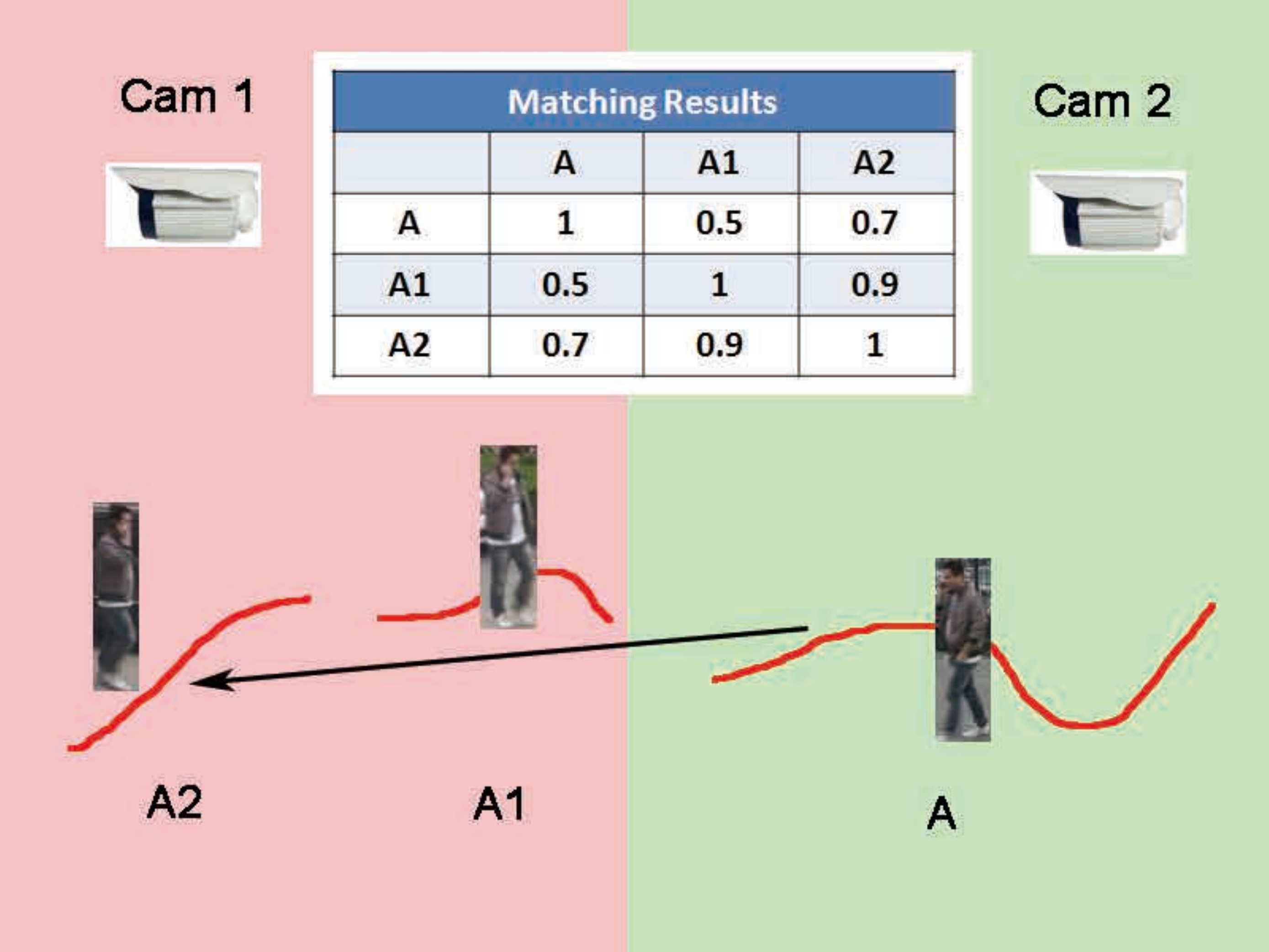}}
  \centerline{(b) Tracklet missing}\medskip
\end{minipage}
\caption{\textbf{Illustration for the two matching problems.} Blue and red lines indicates two targets and arrows show the best matching. Target B is matched to tracklet A2 wrongly in (a). Tracklet A1 is missing in (b).}
\label{fig:trackletproblem}
\end{figure}

As shown in Fig.~\ref{fig:solution} (Solution A), previous visual object tracking approaches tackle
the problem in two different steps:
single camera object tracking (SCT)~\cite{Liu/cvpr2013,Breitenstein/pami2011,Kuo/cvpr2011}
and inter-camera object tracking (ICT)~\cite{Hamid/cvpr2010,Xiaotang,cai2007continuously}.
SCT approaches~\cite{Liu/cvpr2013,Breitenstein/pami2011,Kuo/cvpr2011} attempt to compute the trajectories
of multiple objects from a single camera view,
while ICT approaches~\cite{Hamid/cvpr2010,Xiaotang,cai2007continuously}
aim to find the correspondences among those trajectories across multiple
camera views.
These ICT approaches often use the trajectories obtained from SCT to achieve their data
association, hence the overall tracking system is brittle and the overall performance depends on the results
of the single camera object tracking module.
For challenging scene videos, existing SCT
approaches~\cite{SCT1,SCT2,SCT10move} are also frangible since the results
often contain fragments and false positives.
The direct disturbance of these false positives and fragments bring problems into ICT module,
such as wrong matching problem, \ie two targets in Camera 2 are
matched to different tracklets of a same target in Camera 1 (see
Fig. \ref{fig:trackletproblem} (a)), and tracklet missing problem, \ie some tracklets of a target are missing during inter-camera
tracking (see Fig. \ref{fig:trackletproblem} (b)).
These problems are inevitable as long as the multi-camera object tracking is solved in two steps.
We address these problems by integrating the two separate modules and jointly optimising them.

\IEEEpubidadjcol

We develop a global multi-camera object tracking
approach. It integrates two steps together
via an equalized global graph model to avoid these ``inevitable''
problems and aims to improve the overall performance of multi-camera object tracking.

Considering two different steps, we evaluate the overall performance from the following two criteria:
\begin{itemize}
  \item Single camera object tracking: measuring how well the completed pedestrian trajectories in a single camera can be used to rebuild their exact historical paths in each scene.
  \item Inter-camera object tracking: evaluating how well the inter-camera matching help to locate the pedestrians in a wide area.
\end{itemize}

As shown in Fig. \ref{fig:solution} (Solution A), SCT
and ICT share a similar data association framework: a graph modeling
with an optimisation solution. In the single camera object tracking module, the data
association inputs are the initial observations, such as detections or
tracklets, and the outputs are the integrated trajectories in each
single camera (known as mid-term trajectories). These mid-term
trajectories are then used as inputs to achieve the data association in inter-camera object tracking,
and the outputs of the ICT approaches are the final integrated
trajectories in multi-cameras (known as final trajectories). To
integrate these two data associations, the straightforward idea is to establish a new data association which takes initial observations as inputs and outputs the final trajectories directly.
However, a new problem arises, \ie how to measure the
similarity between two observations in the new graph. Some similarities are from the
observations which belong to the same camera, and others are from those belong to different cameras. If under the
same similarity metric, the average similarity score between observations in
different cameras would be commonly lower\footnote{The higher similarity score indicates a higher likelihood of the link for two observations.} than that from observations in the same camera, because the appearance information and
the spatio-temporal information of objects are less reliable in ICT
than those in SCT due to many factors (camera settings, viewpoints
and lighting conditions). In this case, the optimisation process makes the graph
give priority to linking the observations following the edges in
the same camera instead of those across cameras, which would cause a failed optimized result for the whole multi-camera object tracking. To solve this problem, we have to handle two questions: how to
distinguish the similarities in a same camera from those in
different cameras, and how to balance them in the new graph?
In this paper, we improve the similarity metric, make a difference between similarities of SCT and ICT, and equalize them in a global graph. A minimum uncertain gap~\cite{MUG} is adopted to establish the improved similarity metric. Thanks to this,
the similarity scores in both SCT and ICT are equalized in the proposed global graph model.

The contributions of this paper\footnote{A preliminary version of this paper appeared in
Chen~\etal~\cite{Chen/icip2014} and the source code is available in the link (https://github.com/cwhgn/EGTracker).} are as follows.
\begin{enumerate}
  \item a global graph
model for multi-camera object tracking is presented which integrates SCT and ICT steps together to avoid the ``inevitable'' problems;
  \item an improved
similarity metric is proposed to equalize the different
similarities in two steps and unify them in one graph;
  \item  the
proposed approach is experimented on a comprehensive evaluation
criterion which clearly shows that our method is more effective than the
traditional two-step multi-camera visual tracking framework.
\end{enumerate}
%

\section{Related Work}
\label{sec:relatedwork}

Using a graph model is an efficient and effective way to solve the data
association problem in multi-camera visual object tracking. First,
a graph modeling is used to form a solvable graph model with
input observations (detections, tracklets, trajectories or pairs).
It includes nodes, edges and weights. Then an optimisation solution is
brought in to solve the graph and obtains optimal or suboptimal solutions.
The difference is that single camera object tracking (SCT) emphasizes
particularly on the graph and
the optimisation solution, \ie how to build a more efficient or
more discriminative graph. While inter-camera object tracking (ICT) focuses on nodes, edges and
weights, which prefers getting a more effective feature
representation. The
ICT has more complex and more sophisticated representations
or similarity metrics (\ie a transition matrix), but with a simpler graph model. The
proposed approach takes advantages of both SCT and ICT. The proposed similarity metric is extended from a classical inter-camera tracking method
~\cite{MCSH} and the global graph model takes advantage of a
state-of-the-art SCT approach~\cite{MAP}.

This section introduces related approaches for each
part of SCT, ICT and MCT. Section 2.1 reviews the single camera multi-object
tracking. Section 2.2 discusses the inter-camera
object tracking with a brief introduction of object re-identification.
Section 2.3 shows some other multi-camera object tracking
approaches that take both SCT and ICT into account.

\subsection{Single Camera Object Tracking (SCT)}
\label{ssec:SCT}

In single camera multi-objects tracking, the prediction of the
spatio-temporal information of objects is more reliable and the
appearance of objects does not have many variations during tracking.
This makes the SCT task less challenging than the ICT task. \ie for
some less challenging videos, a simple appearance representation (\textit{e.g.} color histogram
\cite{Chen/cvpr2014,GMCP,Li/cvpr2009}) works well. The graph model is often
used to solve different problems, such
as occlusion \cite{Yang/cvpr2014,Possegger/cvpr2014}, crowd
\cite{Li/cvpr2009,Tang/iccv2013} and interference of appearance
similarity \cite{Dicle/iccv2013,Bae/cvpr2014}. However, for challenging videos,
these approaches lead to frequent id-switch errors and trajectory fragments.

Existing approaches in SCT usually follow a data association-based
tracking framework, which link short tracklets~\cite{Chen/icip2014,GMCP,Wang/cvpr2014}
or detection responses~\cite{Wen/cvpr2014,USCVision1,K-shorts} into
trajectories by a global optimization based on various kinds of features, such
as motion (position, velocity) and appearance (color, shape). The
improvements always develop from two aspects: the graph model and the
optimization solution. Some researchers focus on developing a new graph
model for their tracklets or detections and aim to solve a specific
problem. In Possegger~\etal~\cite{Possegger/cvpr2014}, a geodesic method is
adopted to handle the occlusion problem. Dicle~\etal~\cite{Dicle/iccv2013} use motion dynamics to solve generalized linear
assignments when targets with similar appearances exist. Other works in
SCT focus on the improvement of the optimization
solution framework, such as continuous energy minimization~\cite{CEM},
linear programming~\cite{LP}, CRF~\cite{CRF} and the
mixed integer program \cite{MIP}. Zhang~\etal~\cite{MAP} propose a
maximum a posteriori (MAP) model to solve the data association of
the multi-object tracking, while Yang~\etal~\cite{CRF} utilize
an online CRF approach to handle the optimization with the benefit
of distinguishing spatially close targets with similar appearances.
These approaches can partly yield id-switches and trajectory
fragments, but the separated optimisation makes them suffer from
leaving many fragments and false positives to ICT step.

\subsection{Inter-camera Object Tracking (ICT)}
\label{ssec:ICT}

Inter-camera tracking is more challenging than SCT because of its
greater dramatic changes in appearance caused by many factors
(camera settings, viewpoints and lighting conditions) and less
reliable spatio-temporal information in different camera views. As a
result, how to learn a discriminative and invariant feature representation and
a suitable similarity metric are the main problems in ICT.

Most ICT works solve these problems from multi-camera calibration
\cite{Pflugfelder/avsbs2007,Hu/pami2006,Khan/eccv2006} and feature cues
\cite{USCVision2,Kuo/eccv2010,Matei/cvpr2011,Zheng/icip2011,Zheng/iconip2011}. For multi-camera calibration, as
an immobile information, the approaches in this aspect always
project the multiple scenes into a 3D coordinate system, and achieve
the matching by using projected position information. Hu~\etal~\cite{Hu/pami2006}
adopt a principal axis-based correspondence to achieve the
calibration. For feature cues, most approaches utilize improved appearance or spatio-temporal information
to achieve the matching. Kuo~\etal~\cite{Kuo/eccv2010}
apply a multi-instance learning approach to learn an appearance affinity model,
while Matei~\etal\cite{Matei/cvpr2011} integrate appearance and spatio-temporal
likelihoods within a multi-hypothesis framework.

From the perspective of the graph modeling, a
K-camera ICT data association can be treated as a K-partite graph
matching problem. It is difficult to get the optimal solution, but
there're many approaches to get the suboptimal solutions, \textit{e.g.}
the weighted bipartite graph \cite{Javed/iccv2003}, the Hungarian algorithm~\cite{HungaryAlg}
and the binary integer program~\cite{binaryIP}.
The K-partite idea holds an assumption that each camera has had a
perfect tracking result which should not be changed any more. In
practice, the SCT result is not ideal and the assumption is broken.
In this case, the SCT result should be modifiable and the data
association is more like a global optimization problem than the
K-partite graph matching problem.

At the end of introducing ICT, it is worth mentioning that object re-identification (Re-ID) is an
important part in ICT. When the topology of the camera network is not
available or the scenes are not overlapped, the spatio-temporal
information is invalid. In this case, the appearance cue is the only
information can be used for matching. Studying object
re-identification separately helps
to better understand the capability of object matching by using visual
features alone. Most object re-identification improvements mainly
focus on some certain appearance of objects, such as
color~\cite{MCSH,Zhao/cvpr2014}, shape~\cite{Raftopoulos/cvpr2014,Wang/iccv2007} and
texture~\cite{Hamdoun/icdsc2008}. Recently, Li~\etal~\cite{Li/cvpr2014}
successfully apply CNN on Re-ID to extract an effective feature
representation. However the highest identification rate is
still below 0.3 under benchmarks and the approaches are also
not practical.

As we said, the ICT approaches have a common assumption that the
single camera object tracking results are perfectly done and the
trajectories in single cameras are all true positive and integrated
completely. But until now, they are difficult to be achieved.

\subsection{Multi-camera Object Tracking (MCT)}
\label{ssec:MCT}

A good MCT is the ultimate goal for any
researcher in tracking. Most MCT methods follow the two-step
framework, a SCT algorithm plus an ICT algorithm. In the Multi-Camera
Object Tracking Challenge~\cite{MCTchallenge} in ECCV 2014 visual
surveillance and re-identification workshop, methods of most
participating teams are two-step approaches. The winner USC-Vision
team uses a state-of-the-art SCT method~\cite{USCVision1} and a
state-of-the-art ICT method~\cite{USCVision2}.

Besides two-step approaches, there're some multi-camera object
tracking approaches~\cite{POM,Harry,Leal/cvpr2012,Hofmann/cvpr2013} concentrating on
integrating the processes of SCT and ICT into one global graph as this
paper does. They mainly follow a tracking-by-detection paradigm and
form a global association graph (see Fig. \ref{fig:solution}
(Solution C)). Yu~\etal~\cite{Harry} propose a nonnegative discretization
solution for data association and identify people across different
cameras by face recognition. While for real scenes with objects in a
distant view, faces are too small to be recognized.
Hofmann~\etal~\cite{Hofmann/cvpr2013} use a global min-cost flow graph and connect the
different-view detections through their overlapping locations in a
world coordinate space, which is not suitable for the non-overlapping
camera problem.

In this paper, the proposed method uses
tracklet observations as the inputs instead of object detections, which are more reliable for matching. We consider the multi-camera object tracking as a global tracklet association under
a panoramic view (see Fig.~\ref{fig:solution} (Solution B)). And the
similarities of different tracklets in the global tracklet
association are treated differently according to the cameras they belonging to. This framework provides a new solution for
multi-camera object tracking when the SCT performance is not good
enough for the further ICT process. Its local performance in a
specific camera view may be as fragmentary as that of the
traditional SCT methods, even the inter-camera information may
provide some useful feedbacks for each specific camera. But it
overcomes the new problems emerging in ICT when SCT is not good and
offers a better ICT performance. In practice, a
better ICT has stronger practical significance than SCT. For a video
surveillance system, it's more important to locate the objects in
the whole wide area than a single scene.

\section{Global Graph Model}
\label{sec:GlobalGraphModel}

\begin{figure}[t]
  \centering
  \centerline{\includegraphics[width=1.0\linewidth]{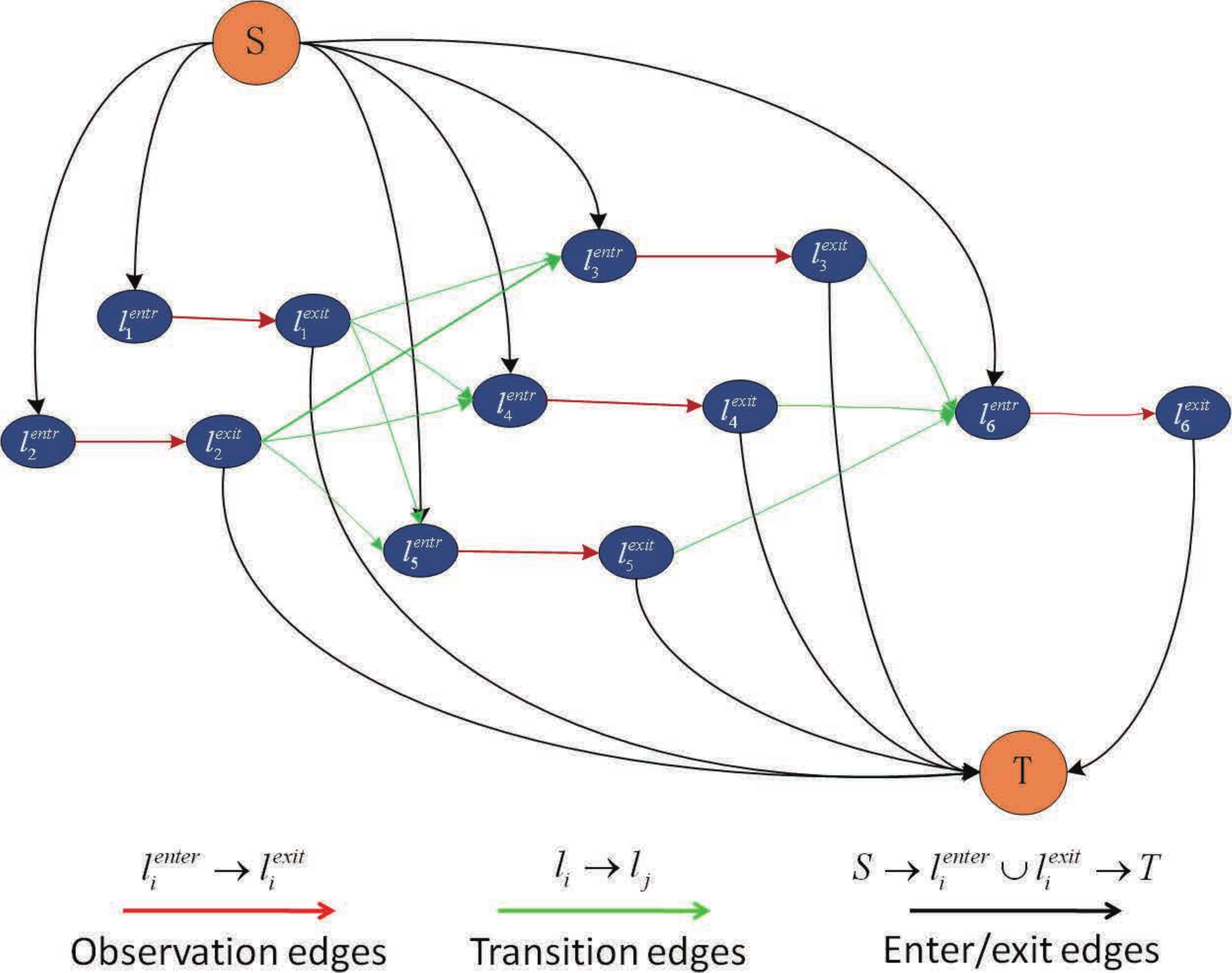}}
%
\caption{\textbf{Illustration for the min-cost flow network.} An example for the min-cost flow network with 3 timesteps and 6 tracklets. The number of $N,E$ and $W$ are 14, 21 and 21.}
\label{fig:MAP}
\end{figure}
Our goal is to predict the trajectories by using the given series of observed videos.
The proposed approach focuses on optimising single camera tracking and inter-camera tracking in
one global data association process. The data association is modeled as a global
maximum a posteriori (MAP) problem which is inspired by the same MAP
formulation from Zhang~\etal~\cite{MAP}. The difference is that the input
in the proposed solution is tracklets rather than object detections.
And the association aims to solve the wrong matching and the tracklet
missing problems in ICT, while Zhang~\etal~\cite{MAP} apply it on SCT.
We outline the variable definitions in Table
\ref{table:MAPNotation}.

In our approach, a single trajectory
hypothesis is defined as an ordered list of target tracklets, i.e.
$\Gamma_i=\{l_{i_1},l_{i_2},...,l_{i_k}\}$ where $l_{i_k}\in L$. The
association trajectory hypothesis $\Gamma$ is defined as a set of
single trajectory hypothesises, i.e. $\Gamma=\{\Gamma_i\}$. The
objective of the data association is to maximize the posteriori
probability of $\Gamma$ given the tracklet set $L$ under the
non-overlapping constraints~\cite{MAP}:

\begin{equation}
\begin{array}{ll}
\Gamma^*=\arg\max\limits_\Gamma\prod\limits_i P(l_i|\Gamma)\prod\limits_{\Gamma_k\in \Gamma}P(\Gamma_k) \\
\Gamma_i\cap \Gamma_j=\emptyset, \forall i\neq j.
\end{array}
\end{equation}

$P(l_i|\Gamma)$ is the likelihood of tracklet $l_i$. The prior
$P(\Gamma_k)$ is modeled as a Markov chain containing transition
probabilities $\prod P(l_{k_{i+1}}|l_{k_i})$ of all tracklets in
$\Gamma_k$ \cite{Hofmann/cvpr2013}.

The transition probability $P(l_j|l_i)$ is computed by using
probabilities of the appearance feature $P_a(l_i\rightarrow l_j)$ and
the motion feature $P_m(l_i\rightarrow l_j)$.

\begin{equation}
P(l_j|l_i)=P(l_i\rightarrow l_j)=(P_a(l_i\rightarrow l_j))^{k_1}\cdot (P_m(l_i\rightarrow l_j))^{k_2},
\end{equation}
where $k_1$ and $k_2$ are the weights of two features.

The MAP association model can be solved by a min-cost flow network
\cite{Chen/icip2014}. The min-cost flow graph is formulated as $G=\{N,E,W\}$,
where $N,E,W$ stands for nodes, edges and weights  respectively and
the weight means the cost of linking the edge. In the graph $G$,
there are two nodes $i^{enter}$ and $i^{exit}$ defined for each
tracklet $l_i$. The observation edge $e_i$ from node $i^{enter}$ to
$i^{exit}$ indicates the likelihood of tracklet $l_i$. The
corresponding observation weight $w_i$ is set to the negative
logarithm of the likelihood $P(l_i|\Gamma)$.

The possible linking relationship between any two tracklets is expressed as a transition edge $e_{ij}$ from node $i^{exit}$ to node $j^{enter}$, the transition weight $w_{ij}$ is the negative logarithm of the transition probability $P(l_j|l_i)$, as shown as follows,
\begin{equation}
w_{i}=-\log \frac{P(l_i|\Gamma)}{1-P(l_i|\Gamma)}.
\end{equation}

The transition weight can also be decomposed into probabilities in continuity of appearance and motion,
\begin{equation}
w_{ij}\!=\!-\!\log P(l_j|l_i)\!=\!-\!k_1\!*\!\log P_a(l_i\!\rightarrow\! l_j)\!-\!k_2\!*\!\log P_m(l_i\!\rightarrow\! l_j).
\end{equation}

In addition to these nodes and edges, there are two extra nodes $S,T$. They are virtual source and sink for the min-cost flow graph. The enter/exit edges $e_{Si}$ and $e_{jT}$ are also added in to represent the start tracklet $l_i$ and the end tracklet $l_j$. The enter/exit weights of these tracklets are both set to 0 in this paper, because every tracklet could be equally a start or end with no cost.

In summary, the number of nodes ($N$) is ($2M+2$), and the numbers of edges $E$ and weights $W$ are smaller than the numbers of full connection graph ($3M+2*{2M \choose 2}$). $M$ is the total number of tracklets in all cameras. As shown in Fig. \ref{fig:MAP}, the graph is solved by the min-cost flow, and the optimal solution is the maximum of the posteriori probability of $\Gamma$ with the minimum cost.

In the rest of this section, we introduce every part of the min-cost flow graph, especially for the weights $W$.

\begin{table}
\footnotesize \caption{Notations of Equalized Global Graph Model}
\label{table:MAPNotation}
\begin{center}
\resizebox{\linewidth}{!}{
\begin{tabular} {ll}
  \hline
    $l_i$ & A single input tracklet consisted of several
    attributes,\\
     & $l_i=[x_i,c_i,s_i,t_i,a_i]$.\\
    $L$ & The set of all input tracklets, $L={l_1,l_2,..,l_M}$.\\
    $\Gamma_i$ & A single trajectory hypothesis consisted of an ordered list of \\
     & target tracklets, $\Gamma_i=\{l_{i_1},l_{i_2},...,l_{i_k}\}$.\\
    $\Gamma^*$ & The output of the aglorithm which is the optimal set of trajectory\\
     & hypothesis.\\
    $G$ & The min-cost flow graph, $G=\{N,E,W\}$.\\
    $N$ & The set of nodes in the graph, $N=\{S,T,l^{enter}_i,l^{exit}_i\}$\\
     & $i\in[1,M]$.\\
    $E$ & The set of edges in the graph,
    $E=\{e_i\}\cup\{e_{Si},e_{iT}\}\cup\{e_{ij}\}$\\
     & $i\in[1,M]$.\\
    $W$ & The set of weights in the graph, $W=\{w_i\}\cup\{w_{Si},w_{iT}\}\cup\{w_{ij}\}$\\
     & $i\in[1,M]$.\\
    $h^i_n$ & The MCSHR of tracklet $l_i$ in the $n$th frame.\\
    $H_i$ & The incremental MCSHR for the whole tracklet $l_i$.\\
    $\Lambda_{k,j}$ & The similarity between any MCSHR pair $h_k$ and
    $h_j$.\\
    $\tau_i$ & The best periodic time for tracklet $l_i$.\\
  \hline
\end{tabular}}
\end{center}
\end{table}

\subsection{Nodes}
\label{ssec:nodes}

In the proposed approach, the tracklets extracted by a single-object tracking method are treated as input observations instead of detections. In other words, these tracklets are used to produce nodes in the global graph model. One of the reasons is that they have more information (like motion) than detections which only contain appearance information. With more information, they can be considered as more credible nodes and the similarities of them are more reliable. What's more, the number of the tracklets is much smaller than that of detections. It's a good way to speed up the computing time of the graph optimization, which is also very important for practical usages. In this paper, the deformable part-based model (DPM) detector~\cite{Felzenszwalb/pami2010} and an AIF tracker~\cite{AIF} are first used to get all the tracklets from each camera. After obtaining detections by the DPM detector, we use the AIF tracker to track every target and get their tracklets. During the target tracking by the AIF tracker, a confidence $\alpha_t$~\cite{AIF} is calculated to evaluate the accuracy of a tracking result in frame $t$. If the confidence score is lower than the threshold $\theta$, \ie $\alpha_t<\theta$, the tracker is considered to be lost. Then all confidence values of the target in previous frames are recorded and the average value $c$ is computed as the likelihood $P(l_i|\Gamma)$ of tracklet $l_i$,

\begin{equation}
c_i=P(l_i|\Gamma)=\frac{\Sigma_{k=t^{start}_i}^{t^{end}_i}\alpha_k}{(t^{end}_i-t^{start}_i)},
\label{eq:AIFconfidence}
\end{equation}
where $t^{start}_i$ and $t^{end}_i$ are the start and end frames of tracklet $l_i$.

So all the tracklets from all cameras are obtained, $L=\{l_1,l_2,...,l_M\}$ , where each tracklet $l_i=[x_i,c_i,s_i,t_i,a_i]$ consists of position, likelihood, camera view, time stamp and appearance information respectively. The nodes $N$ can be expressed as:

\begin{equation}
N=\{S,T,l^{enter}_i,l^{exit}_i\}\phantom{X}i\in[1,M]
\end{equation}

\subsection{Edges}
\label{ssec:edges}

Edges are also an important part for the graph model. All
the observation edges and enter/exit edges are reserved in the
min-cost flow graph. However, for the transition edges, only a part
of it is retained because that not all the edges are meaningful.
Three rules are built for selecting transition edges in our graph.

Firstly, for edge $e_{ij}$, the start frame $t^s_j$ of the tracklet $l_j$
must be after the end frame $t^e_i$ of the tracklet $l_i$ without any
overlapping frame. This rule ensures the uniqueness of objects in every frame and keeps the edges directed.
Secondly, the two tracklets $l_i$ and $l_j$ should come
from the same camera or two cameras with an existing topological
connection, which ensures the link of two tracklets possible from a
panoramic view. Thirdly, a waiting time threshold $\eta$ is brought in
to limit the link of two tracklets. If the time interval between two
tracklets is long enough, longer than the threshold $\eta$, the likelihood
of this link is close to zero. As a result, the
edges that meet all requirements are selected and reserved,

\begin{equation}
\begin{array}{lll}
E=\{e_i\}\cup\{e_{Si},e_{iT}\}\cup\{e_{ij}\}\phantom{X}i\in[1,M],\\
\phantom{XXXXX}0<t_j^{start}-t_i^{end}<\eta,\\
\phantom{XXXXX}Topo(s_i,s_j)=1,
\end{array}
\end{equation}
where $Topo(s_i,s_j)=1$ means the camera views of $s_i$ and $s_j$ have an existing topological connection.

For all these selected edges $E$, the capacity is set to 0 or 1, because every target should be at one and only
one place in the same time. If the capacity is 1 in the optimal
solution, which means this link exists and the two tracklets of this link belong to the same target.

\subsection{Weights}
\label{ssec:weights}

Weights are an essential attribution for links and used to represent relationships between nodes. In this paper, we import the similarities among tracklets as weights to indicate the cost of building links. As it mentioned above, the weights $W$ are consisted of three parts, the same as edges:

\begin{equation}
W=\{w_i\}\cup\{w_{Si},w_{iT}\}\cup\{w_{ij}\}\phantom{X}i\in[1,M]\\
\end{equation}

The observation weights can be obtained according to Eq.~\ref{eq:AIFconfidence}. And the enter/exit weights are all set to 0 as mentioned above. In the transition weights, the appearance similarity $P_a(l_i\rightarrow l_j)$ and the motion similarity $P_m(l_i\rightarrow l_j)$ are used to form the weights. In the following we introduce them respectively.

\begin{equation}
\begin{array}{ll}
\left\{ \begin{array}{llll}
w_i=-\log \frac{P(l_i|\Gamma)}{1-P(l_i|\Gamma)}=-\log \frac{c_i}{1-c_i},\\
w_{Si}=w_{iT}=0 \phantom{XXXXXXXXXXX} i,j\in[1,M],\\
w_{ij}=-\log P(l_j|l_i)\\
\phantom{w_{ij}}=-k_1*\log P_a(l_i\rightarrow l_j)-k_2*\log P_m(l_i\rightarrow l_j).
\end{array} \right.
\end{array}
\end{equation}

\begin{figure}[!t]
  \centering
  \centerline{\includegraphics[width=1.0\linewidth]{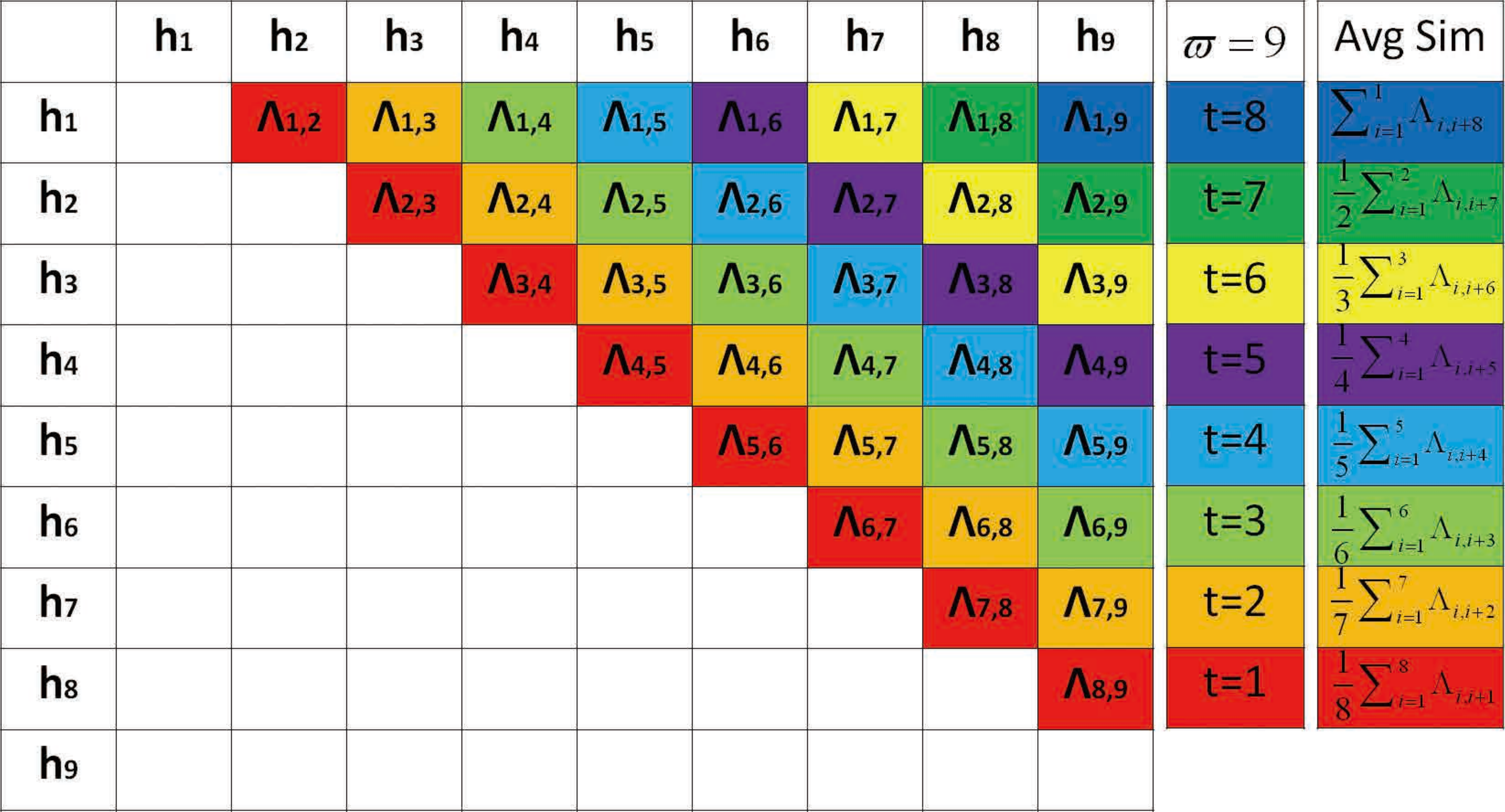}}
%
\caption{\textbf{Illustration of computing the periodic time for a tracklet.} An example for a tracklet with the length $\varpi$ of 9 frames. The Avg Sim column shows the validity of every possible periodic time $t$. The maximum in Avg Sim column indicates the best periodic time $\tau$ for this tracklet.}
\label{fig:example4PMCSHR}
\end{figure}

\subsubsection{Appearance Similarity}
\label{sssec:asim}

As shown in Section \ref{sec:relatedwork}, both SCT and ICT have
their own representations and similarity metrics, while those in ICT
methods are more sophisticated than those in SCT ones. In order to build an equalized
metric, the proposed approach adopts an ICT representation. But it doesn't use any learning process which strongly increases the computing time. This representation is called
Piecewise Major Color Spectrum Histogram Representation (PMCSHR)
\cite{Chen/icip2014}. It's an improved version of Major Color Spectrum
Histogram Representation (MCSHR)~\cite{MCSH} with some
periodicity information that is specific to pedestrian. MCSHR
obtains the major colors of a target based on an online k-means
clustering algorithm. The original way of computing the MCSHR of a tracklet is to integrate histograms in all frames
together.

\begin{equation}
H_i=\frac{1}{\varpi_i}\sum_{n=1}^{\varpi_i} h^i_n,
\end{equation}
where $h^i_n$ is the MCSHR of tracklet $l_i$ in the $n$th frame and
$H_i$ is the incremental MCSHR~\cite{MCSH} for the whole tracklet
$l_i$. $\varpi_i$ is the length of tracklet $l_i$.

As non-rigid targets, pedestrians are challenging objects to be tracked even with the help of the MCSHR.
However, we can make some assumptions to help tracking. We assume that pedestrians are always walking at a constant speed in scenes, and the goal of our approach is to find the periodic time $\tau_i$ to segment the tracklets.

All MCSHRs $\{h_1,h_2,...,h_{\varpi_i}\}$ of the
tracklet $l_i$ are firstly obtained, and then the similarity $\Lambda_{k,j}$
between any pair $h_k$ and $h_j$ is computed. The intuition is to compute
all the possible periodic times and find the best one. For a certain
periodic time $t$, the similarity $\Lambda_{j,j+t}$ between $h_j$
and its next periodic $h_{j+t}$ is collected for every frame $j$, and the average similarity is considered as the value
which determines the validity of this periodic time $t$. As shown in Fig. \ref{fig:example4PMCSHR}, the periodic time with a
highest validity is considered as our best periodic time $\tau_i$
for tracklet $l_i$.

\begin{equation}
\tau_i=\arg\max_t\frac{1}{\varpi_i-t}\sum_{j=1}^{\varpi_i-t}\Lambda_{j,j+t} \phantom{X}\forall t\in [\gamma,\varpi_i/2).
\label{eq:period}
\end{equation}

The set $[\gamma,\varpi_i/2)$ is used to limit the possible range of
$t$, and $\gamma$ is set to 15. If $\gamma$ is too small, the nearby
frames will have a strong similarity which causes Eq.~\ref{eq:period} to
a false maximum. After calculation, $\tau_i$ is the best periodic
time for tracklet $l_i$. Then the tracklet $l_i$ can be evenly
segmented into pieces with the length $\tau_i$ (except the end
part). For each piece, the incremental MCSHR is computed. The PMCSHR
of tracklet $l_i$ is represented by $\{H^i_1,H^i_2,...,H^i_{d_i}\}$,
and $d_i=\lceil \frac{\varpi_i}{\tau_i}\rceil$ is the number of
pieces that the tracklet $l_i$ is segmented into.

Then every similarity between each two pieces from tracklets $l_i$
and $l_j$ are computed, and the average similarity $Dis(l_i,l_j)$ is
considered as the appearance similarity between two tracklets.

\begin{equation}
P_a(l_i\rightarrow l_j)=Dis(l_i,l_j)=\frac{1}{d_i*d_j}\sum_{n=1,m=1}^{d_i,d_j}Sim(H^i_n,H^j_m),
\label{eq:appearance_similarity_s}
\end{equation}
where $Sim(H^i_n,H^j_m)$ is the similarity metric for two tracklets' incremental MCSHRs.

\begin{figure}[!t]
  \centering
  \centerline{\includegraphics[width=1.0\linewidth]{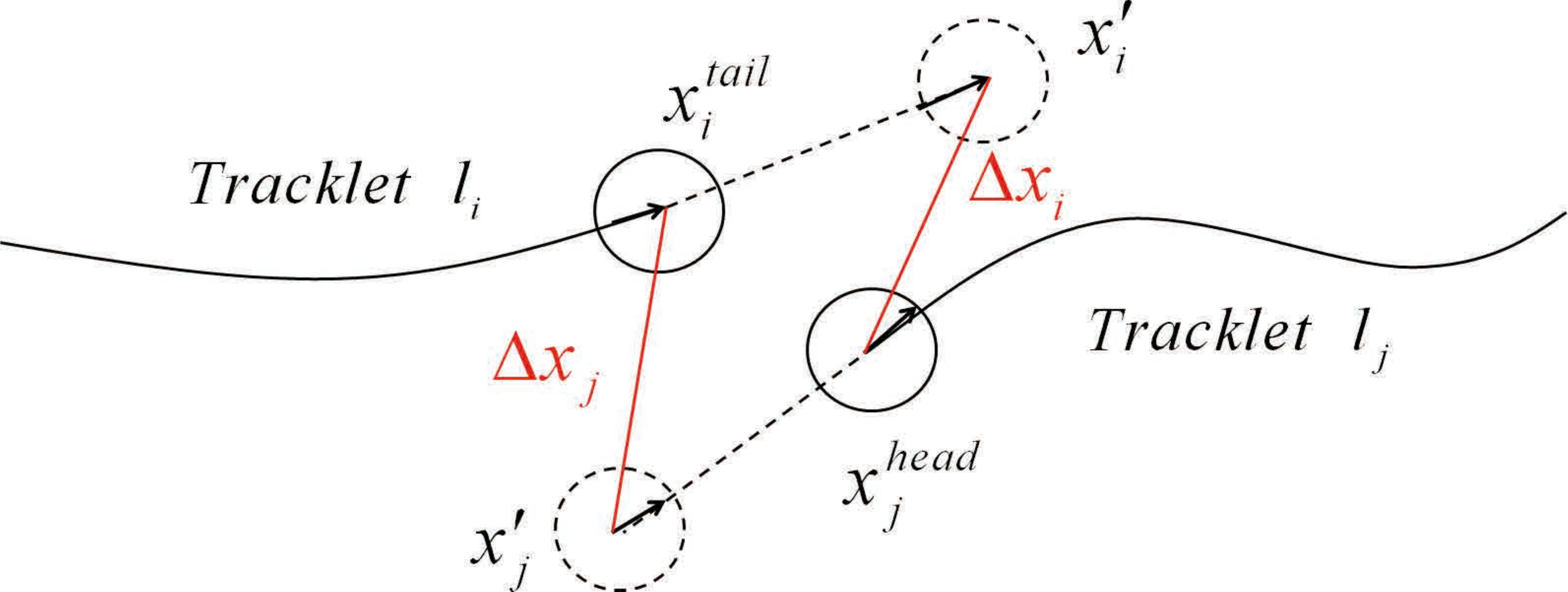}}
%
\caption{\textbf{Illustration of the calculation of the relative distance.}}
\label{fig:motion_s1}
\end{figure}

\begin{figure}[!t]
\begin{minipage}[b]{0.45\linewidth}
  \centering
  \centerline{\includegraphics[width=1.0\linewidth]{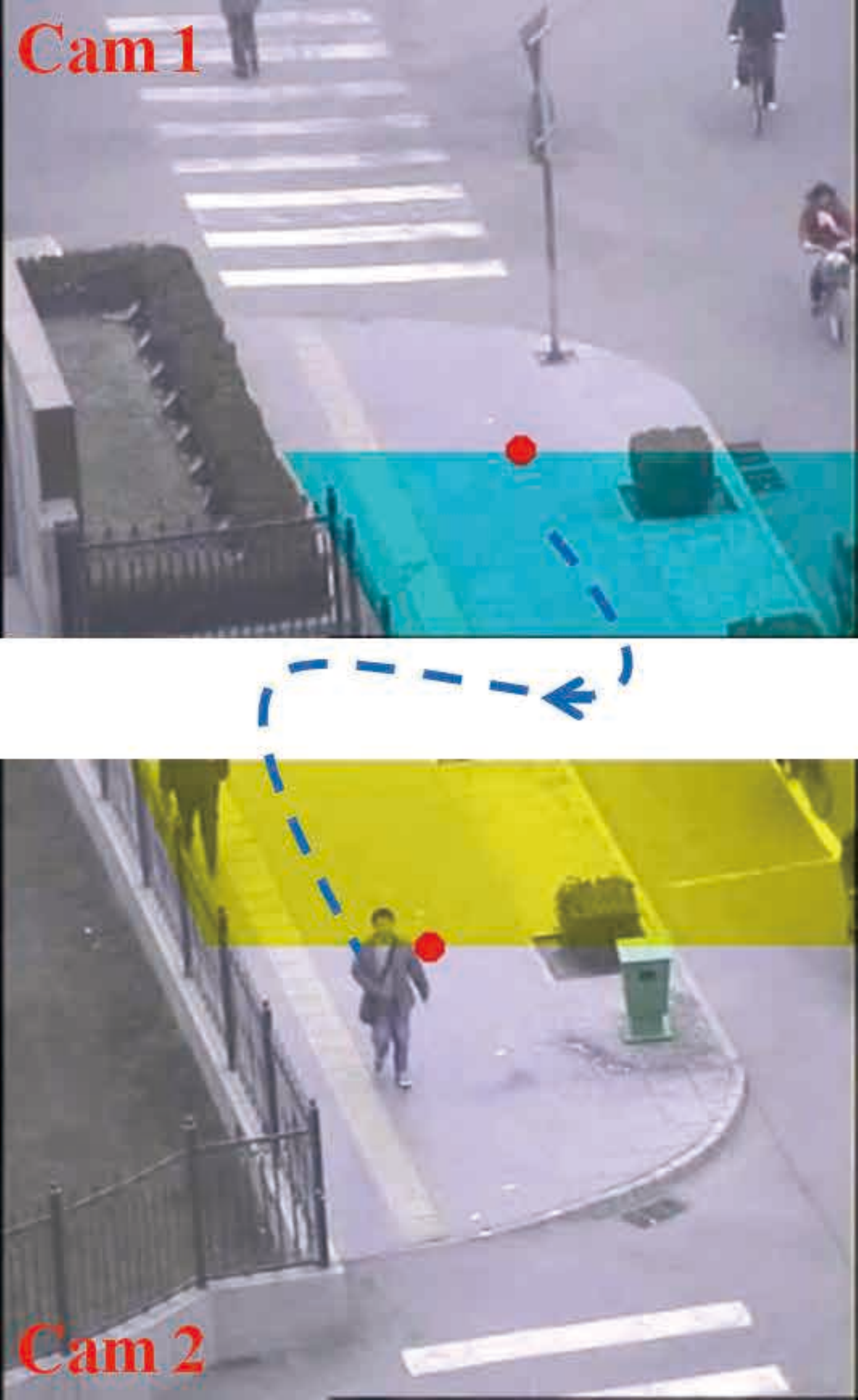}}
  (a)
\end{minipage}
\hfill
\begin{minipage}[b]{0.45\linewidth}
  \centering
  \centerline{\includegraphics[width=1.0\linewidth]{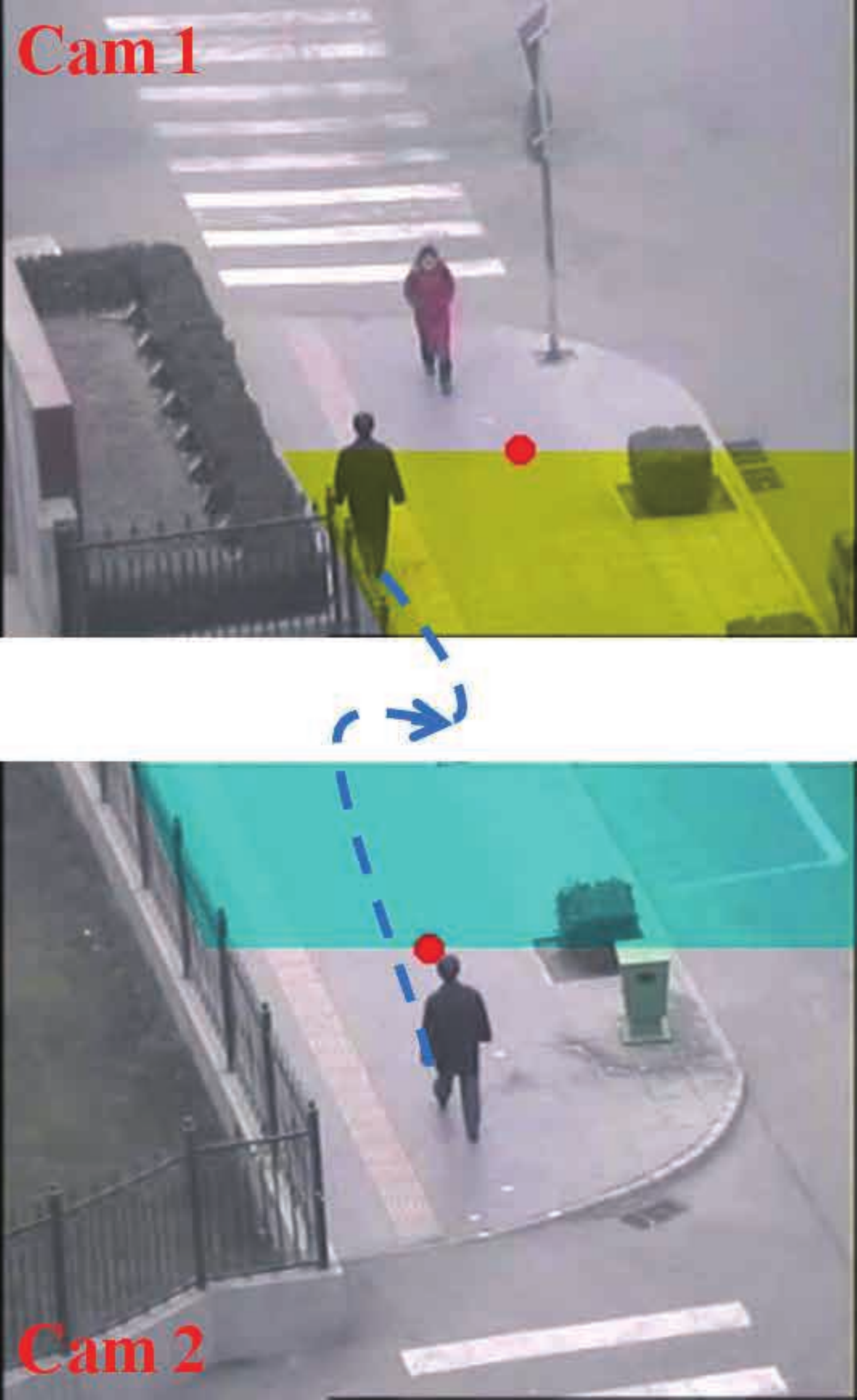}}
  (b)
\end{minipage}
\caption{\textbf{Illustration of the enter/exit areas for the multi-camera visual object tracking.} The enter/exit areas for links from Cam 1 to Cam 2 are in column (a), while those from Cam 2 back to Cam 1 are in column (b). The blue and yellow areas indicates the exit and enter areas respectively, and the red points represent the disappearing points.}
\label{fig:motion_s2}
\end{figure}

\subsubsection{Motion Similarity}
\label{sssec:msim}

For a general method that is available in both overlapping and non-overlapping views, it's hard to always build an exact 3D coordinate system to project all scenes together. Hence, in this paper a relative distance between two tracklets is adopted to measure the motion similarity. For two tracklets $l_i$ and $l_j$, it's easy to get their interval time by a simple subtraction. If the two tracklets are likely to belong to one target, the interval time $t_{ij}^{inv}$ must be a positive number.

\begin{equation}
t_{ij}^{inv}=t_j^{start}-t_i^{end},
\end{equation}
where $t_j^{start}$ is the start time of tracklet $l_j$ and $t_i^{end}$ is the end time of tracklet $l_i$.

With the interval time $t_{ij}^{inv}$, the position $x_i^{tail}$ and the velocity $v_i^{tail}$ of tracklet $l_i$ in the end time, we can predict the position where the tracklet $l_i$ is behind $t_{ij}^{inv}$ time. The new position can be calculated as below:

\begin{equation}
x'_i=x_i^{tail}+v_i^{tail}*t_{ij}^{inv}.
\end{equation}

For tracklet $l_j$, we can conduct the same thing and get its predicted position $t_{ij}^{inv}$ time ago.

\begin{equation}
x'_j=x_j^{head}-v_j^{head}*t_{ij}^{inv}.
\end{equation}

As people always walk along a smooth path in real scene, we can assume that if the two tracklets belong to a same person, the corresponding predicted positions must be close to each other. In other words, $x'_i$ and $x'_j$ should be close enough to $x_j^{head}$ and $x_i^{end}$ respectively. Therefore, the distances between predicted positions and original positions are used to represent the motion similarity between two tracklets (seen in Fig. \ref{fig:motion_s1}).

So the motion similarity in the single camera is computed as below:

\begin{equation}
P_m(l_i\rightarrow l_j)=exp(-\frac{\lambda}{2}(\Delta{x_i}+\Delta{x_j}))\phantom{X}s_i=s_j.
\label{eq:motion_similarity_s}
\end{equation}

\begin{figure}[!t]
\begin{minipage}[b]{0.3\linewidth}
  \centering
  \includegraphics[width=1.0\linewidth]{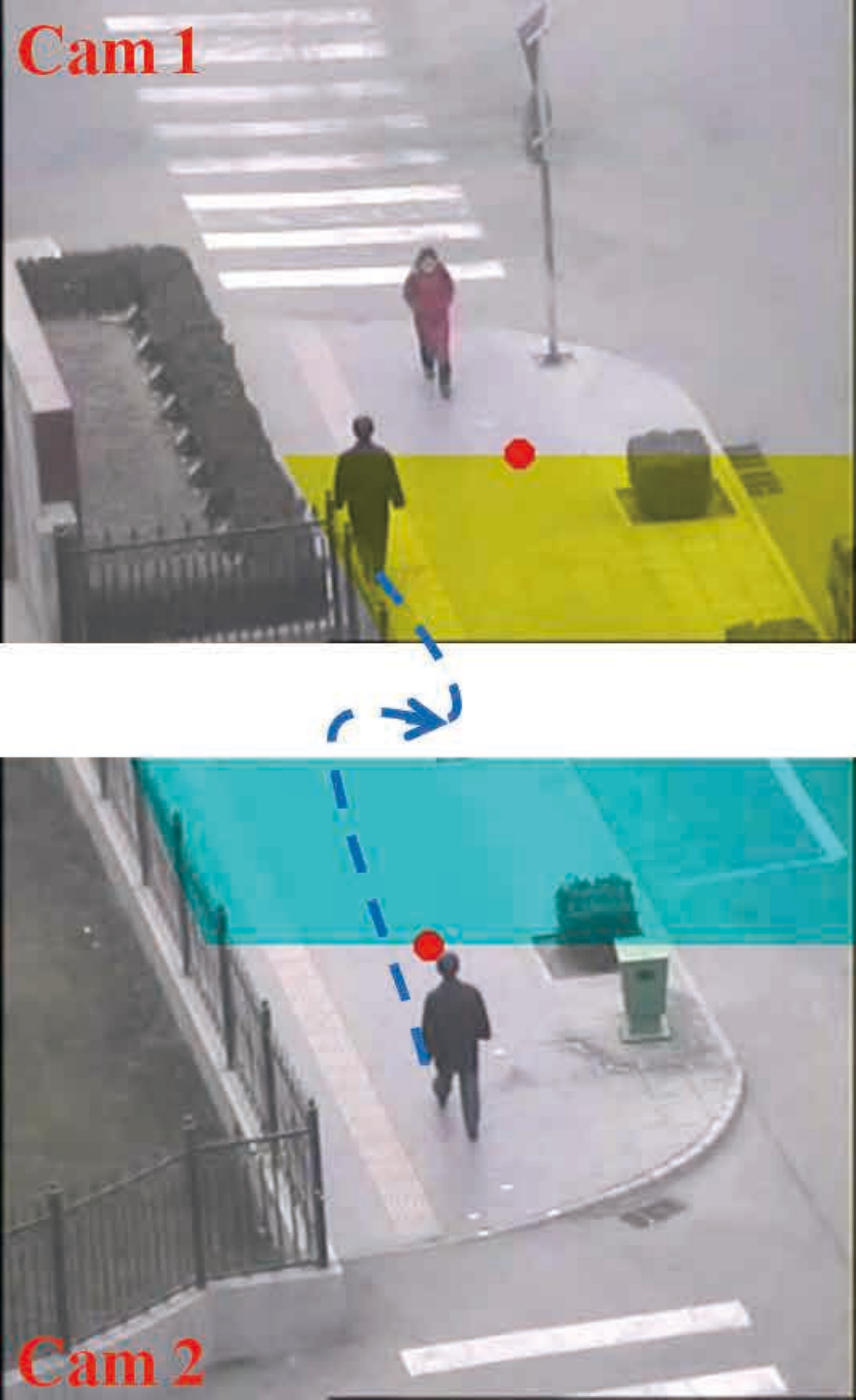}
  (a)
\end{minipage}
\hfill
\begin{minipage}[b]{0.3\linewidth}
  \centering
  \includegraphics[width=1.0\linewidth]{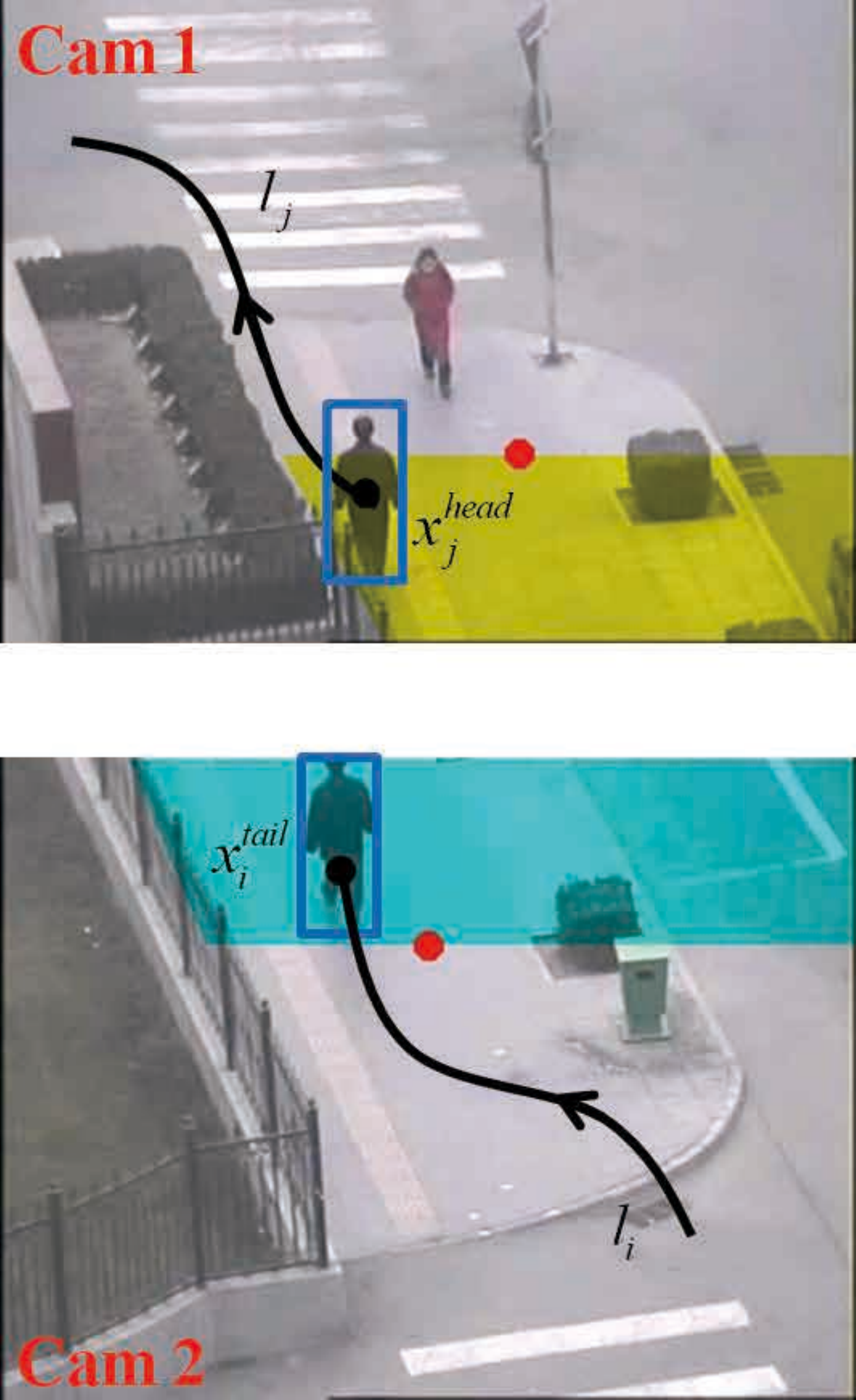}
  (b)
\end{minipage}
\hfill
\begin{minipage}[b]{0.3\linewidth}
  \centering
  \includegraphics[width=1.0\linewidth]{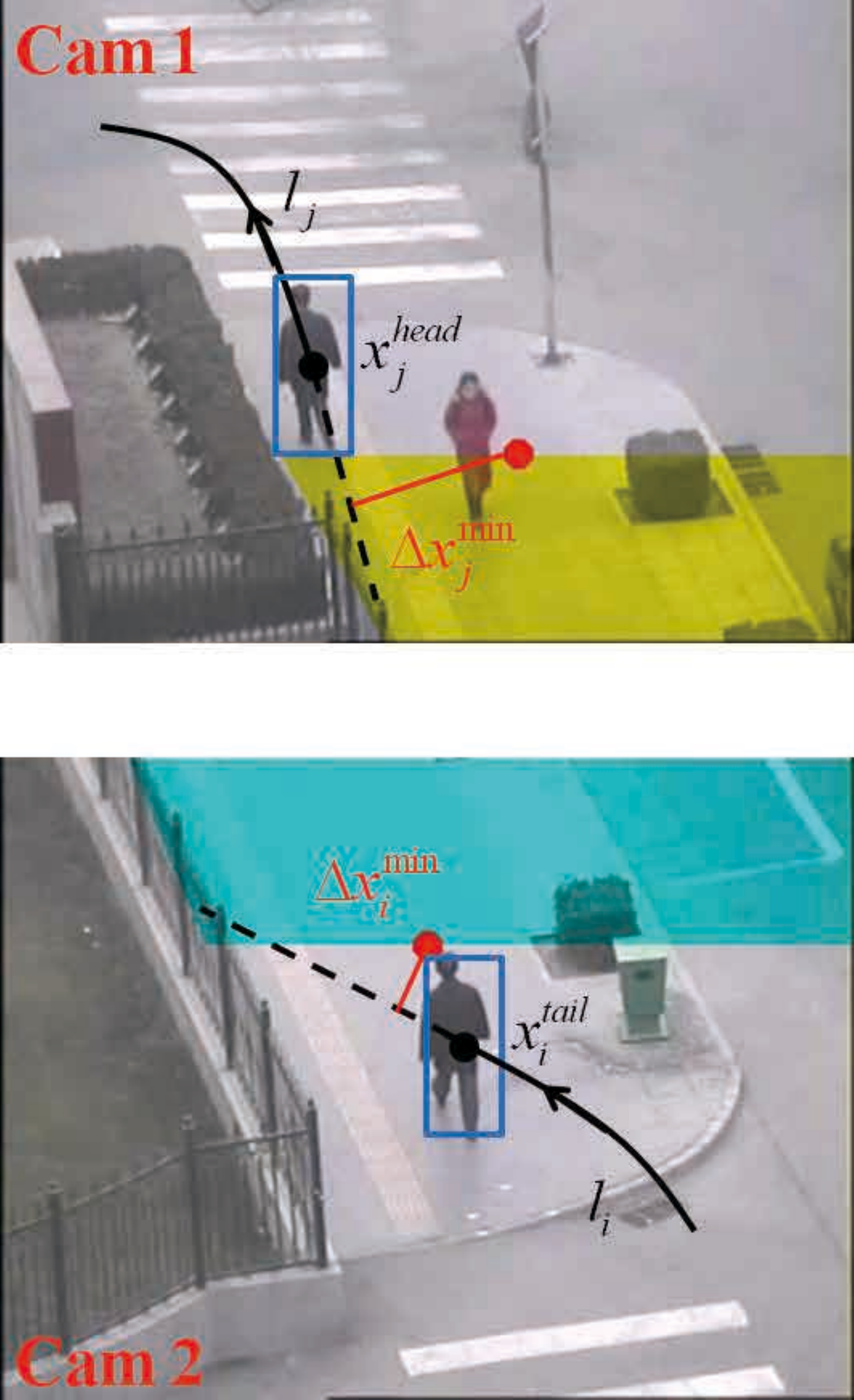}
  (c)
\end{minipage}
\caption{\textbf{Illustration of the computing method for the minimum relative distance across cameras.} In column (b), $x^{tail}_i$ and $x^{head}_j$ are in exit and enter areas respectively, which indicate that both of $\Delta{x^{min}_i}$ and $\Delta{x^{min}_j}$ are set to 0. The red lines in column (c) are $\Delta{x^{min}_i}$ and $\Delta{x^{min}_j}$.}
\label{fig:motion_c}
\end{figure}

As shown in Eq.~\ref{eq:motion_similarity_s}, the
relative distance is only valid for two tracklets from the same
camera. If tracklets are from different cameras, the interval time
is partly invalid. Becasue in inter-camera cases, the pathes between
cameras are hard to measure which renders the interval time useless
for predicting positions. In this case, the relative distance mostly
tends to be a huge wrong number. To handle this problem, a minimum
relative distance is applied to compute the similarity across
cameras, which is comparable with Eq.~\ref{eq:motion_similarity_s}.

Enter/exit areas are commonly used in some uncalibrated camera systems to help to re-local exact positions of targets. Hence, we labeled enter/exit areas of each camera view with the help of topology information (seen Fig. \ref{fig:motion_s2}).

For a person, if she disappeared from an exit area, we would assume that she could be found in the enter area of the possible corresponding camera (seen in Fig. \ref{fig:motion_c} (a)). If she disappeared from an area near a exit area, she could re-appear in the possible corresponding enter area with a high probability. Under this assumption, we manually set a disappearing point for each area to connect cameras. Then a minimum relative distance $\Delta{x^{min}_i}$ to the disappearing point during the whole interval time is adopted to measure the motion similarity across cameras instead of the original relative distance $\Delta{x_i}$, seen in Fig. \ref{fig:motion_c} (b) and (c).

\begin{equation}
\begin{array}{l}
\Delta{x^{min}_i} \!=\! \left\{ \begin{array}{llll}
\min\limits_{t\in[1,t^{inv}]}\|x_i^{tail}\!+\!v_i^{tail}\!*\!t\!-\!x_{s_i}^{exit}\|_2, \\
\phantom{XXXXXXXXXX}\textrm{if $x_i^{tail}\notin Area_{exit}$},\\
0\\
\phantom{XXXXXXXXXX}\textrm{if $x_i^{tail}\in Area_{exit}$}.
\end{array} \right.
\end{array}
\end{equation}

\begin{equation}
\begin{array}{l}
\Delta{x^{min}_j} \!=\! \left\{ \begin{array}{llll}
\min\limits_{t\in[1,t^{inv}]}\|x_j^{head}\!-\!v_j^{head}\!*\!t\!-\!x_{s_j}^{enter}\|_2 \\
\phantom{XXXXXXXXXX} \textrm{if $x_i^{head}\notin {Area}_{{enter}}$}\\
0 \\
\phantom{XXXXXXXXXX} \textrm{if $x_i^{head}\in {Area}_{{enter}}$}.
\end{array} \right.
\end{array}
\end{equation}

\begin{equation}
P_m(l_i\rightarrow l_j)=exp(-\frac{\lambda}{2}(\Delta{x^{min}_i}+\Delta{x^{min}_j})),
\label{eq:motion_similarity_c}
\end{equation}
where $x_{s_i}^{exit}$ and $x_{s_i}^{enter}$ are the positions of the disappearing points for the enter area and the exit area in camera $s_i$ respectively.

Another benefit of the minimum relative distance is that it is measured in each camera which can be compared with the relative distance. With its help, the motion similarity metric can be extend from a single camera to a multi-camera system and can be considered as well equalized in the global graph.

The final equalized motion similarity metric is:

\begin{equation}
P_m(l_i\rightarrow l_j) = \left\{ \begin{array}{ll}
exp(-\frac{\lambda}{2}(\Delta{x_i}+\Delta{x_j})) & \textrm{if $s_i=s_j$}\\
exp(-\frac{\lambda}{2}(\Delta{x^{min}_i}+\Delta{x^{min}_j})) & \textrm{if $s_i\neq s_j$},
\end{array} \right.
\end{equation}
where $\lambda$ is set to 0.01 in the experiments.

\section{Equalized Graph Model}
\label{sec:EqualizedGraphModel}

During tracking objects in a single camera, we assume that observations
are obtained under the same circumstance, like illumination and
angle of view. Hence the targets would have a strong invariance in their
appearance representations which can further be used for tracking. During
inter-camera object tracking, this invariance is weaker due to
the changes in different circumstances. When we establish the graph
with nodes and edges, this phenomenon would cause the inter-camera
similarities being much lower than the similarities in single camera.
If we use Eq. \ref{eq:appearance_similarity_s} to compute the appearance similarities
and provide no alignment or equalization for two similarity
distributions, it
would result in that the optimization process links the edges in the single camera
preferentially all the time and ignores the inter-camera links as
long as there is a edge with a higher similarity in the same camera. It's hard to get an accurate
alignment for two similarity distributions, and the proposed approach
offers a suitable alignment which can be considered as a
compensation for the inter-camera similarities. Our purpose is to
equalize the difference between two similarity distributions and at
the same time manage to keep the distribution of the inter-camera
similarity not affected. So our equalization is mainly processed on the
distribution of the single camera similarity and make it close to the inter-camera similarity distribution.

\begin{equation}
\begin{array}{ll}
P_a(l_i\rightarrow l_j)=\Delta\sigma(Dis(l_i,l_j)-\Delta\mu), \\
\phantom{XXXXX}\Delta\mu\geq0,s_i=s_j,\\
\end{array}
\end{equation}
where $\Delta\sigma$ and $\Delta\mu$ are the compensation factors, the similarity $Dis(l_i,l_j)$ between tracklets $l_i$ and $l_j$ is obtained by Eq. \ref{eq:appearance_similarity_s}.

The factor $\Delta\mu$ is used to improve the average level of the single camera similarity distribution and the factor $\Delta\mu$ is adopted to control the amplitude of variation. They are computed from two similarity distributions.

\begin{equation}
\begin{array}{ll}
\Delta\mu=\mu_1-\mu_2, \\
\Delta\sigma=\sigma_2/\sigma_1,
\end{array}
\end{equation}
where $\mu_1$ and $\sigma_1$ are the mean and variance of the single camera similarity distribution. These should be computed by all the single camera edges. And $\mu_2$ and $\sigma_2$ are of the inter-camera similarity distribution and should be got from all the inter-camera edges.

However, not all the similarities of edges are reliable and suitable to compute the mean and variance. Some have a large proportion of noises and should be excluded as outliers. In this paper, a minimum uncertain gap (MUG)~\cite{MUG} is brought in to help to filtrate edges used for computing the mean and variance. The MUG is used to measure the uncertainties of the likelihoods between tracklets. The tracklet link with a small MUG can be considered as a more reliable link, because its similarity is more stable and more believable. As a result, the MUG is treated as a confidence factor for edges.

\begin{equation}
\begin{array}{ll}
MUG(l_i,l_j)=\max Sim(H^i_n,H^j_m)-\min Sim(H^i_n,H^j_m) ,\\
\phantom{XXXXX}n\in[1,d_i],m\in[1,d_j].
\end{array}
\end{equation}

Therefore, with the help of MUG's filtration, the mean and variance are computed as follows:

\begin{equation}
\begin{array}{ll}
\mu_1= MEAN(Dis(l_i,l_j)) \phantom{X} \sigma_1= VAR(Dis(l_i,l_j)) ,\\
\phantom{XXXXX}MUG(l_i,l_j)<\varepsilon,s_i=s_j.
\end{array}
\end{equation}

\begin{equation}
\begin{array}{ll}
\mu_2= MEAN(Dis(l_i,l_j)) \phantom{X} \sigma_2= VAR(Dis(l_i,l_j)) ,\\
\phantom{XXXXX}MUG(l_i,l_j)<\varepsilon,s_i\neq s_j,
\end{array}
\end{equation}
where $\varepsilon$ is a confidence threshold, MEAN() and VAR() are the mean and variance operations respectively.

And the final equalized appearance similarity metric would become:

\begin{equation}
P_a(l_i\rightarrow l_j) = \left\{ \begin{array}{ll}
Dis(l_i,l_j) & \textrm{if $s_i\neq s_j$},\\
\Delta\sigma(Dis(l_i,l_j)-\Delta\mu) & \textrm{if $s_i=s_j$}.
\end{array} \right.
\end{equation}

\section{Experiment Results}
\label{sec:ExperimentResults}

In this section, the proposed approach is evaluated based on the
following aspects. First, the global graph model is compared with the
traditional two-step framework, where we use the same feature
representation for fairness. Second, a performance comparison
between the equalized graph and the non-equalized one is provided to prove the
effectiveness of the equalization process with the improved similarity
metric. Third, the proposed approach is compared with some
state-of-the-art Multi-Camera Tracking (MCT) methods. However, as there're no benchmark for
MCT, we introduce a dataset and a comprehensive evaluation
criterion first, which can be developed as a benchmark in further
works. The dataset is specialized for multi-camera pedestrian
tracking in non-overlapping cameras, called NLPR$\_$MCT
dataset. The details of the dataset are presented in
Section~\ref{ssec:datasets}. The proposed evaluation criteria for MCT is introduced
in Section~\ref{ssec:criteria}.

\begin{figure*}[!t]
  \centering
  \centerline{\includegraphics[width=0.95\linewidth]{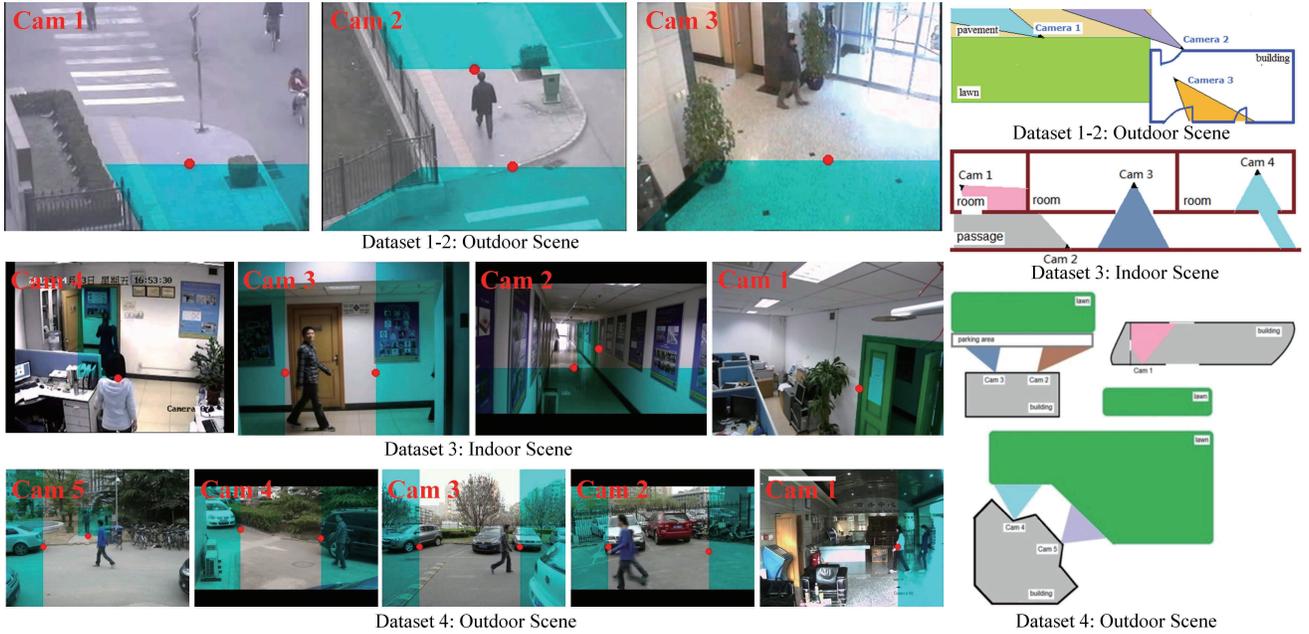}}
%
\caption{\textbf{Illustration of the topological relationship during tracking.} The topological relationships for every dataset are shown in the right column, and the blue polygons stand for enter/exit areas used in our experiments for Dataset 1-4.}
\label{fig:scene}
\end{figure*}

\begin{table*}
\renewcommand{\arraystretch}{1.5}
\caption{The single-camera and inter-camera ground truthes
for all four sub-datasets.}
\begin{center}
\resizebox{\linewidth}{!}{
\begin{tabular} {|c|c|c|c|c|c|c|c|}
  \hline
    \multicolumn{2}{|c|}{Dataset1} & \multicolumn{2}{c|}{Dataset2} & \multicolumn{2}{c|}{Dataset3} & \multicolumn{2}{c|}{Dataset4} \\
  \hline
    $Tracking^{SCT}$ & $Tracking^{ICT}$ & $Tracking^{SCT}$ & $Tracking^{ICT}$ & $Tracking^{SCT}$ & $Tracking^{ICT}$ & $Tracking^{SCT}$ & $Tracking^{ICT}$\\
  \hline
    71853 & 334 & 88419 & 408 & 18187 & 152 & 42615 & 256 \\
  \hline
\end{tabular}}
\end{center}
\label{table:Gth}
\end{table*}

\subsection{Datasets}
\label{ssec:datasets}

For a comprehensive performance evaluation, it is crucial to develop
a representative dataset. There are several datasets for visual
tracking in the surveillance scenarios, such as
PETS~\cite{PETS2009}, CAVIAR~\cite{CAVIAR}, TUD~\cite{TUD} and
i-LIDS~\cite{iLIDS} databases. However, most of them are designed for
multi-object tracking in a single camera and are not suitable for
inter-camera object tracking.
PETS is under a simulation environment with overlapping cameras, not
in real scene, while i-LIDS aims to serve multi-camera object tracking
indoor and the ground truthes are not for free so far. For these
reasons, a new pedestrian dataset is constructed in this paper for
multi-camera object tracking to facilitate the tracking evaluation.

The NLPR$\_$MCT dataset\footnote{http://mct.idealtest.org/Datasets.html} consists of four
sub-datasets. Each sub-dataset includes 3-5 cameras with
non-overlapping scenes and has a different situation according to
the number of people (ranging from 14 to 255) and the level of
illumination changes and occlusions. The collected videos contain
both real scenes and simulation environments. We also list the
topological connection matrixes for pedestrian walking areas. All the videos
are nearly 20 minutes (except Dataset 3) with a rate of 25 fps and are
recorded under non-overlapping views during daily
time, which make the dataset a good representation of different
situations in normal life. The connection relationships between
scenes are shown in Fig.~\ref{fig:scene}, where the enter/exit areas
for this paper are also marked.

%

\begin{figure*}[t]
  \centering
  \centerline{\includegraphics[width=0.95\linewidth]{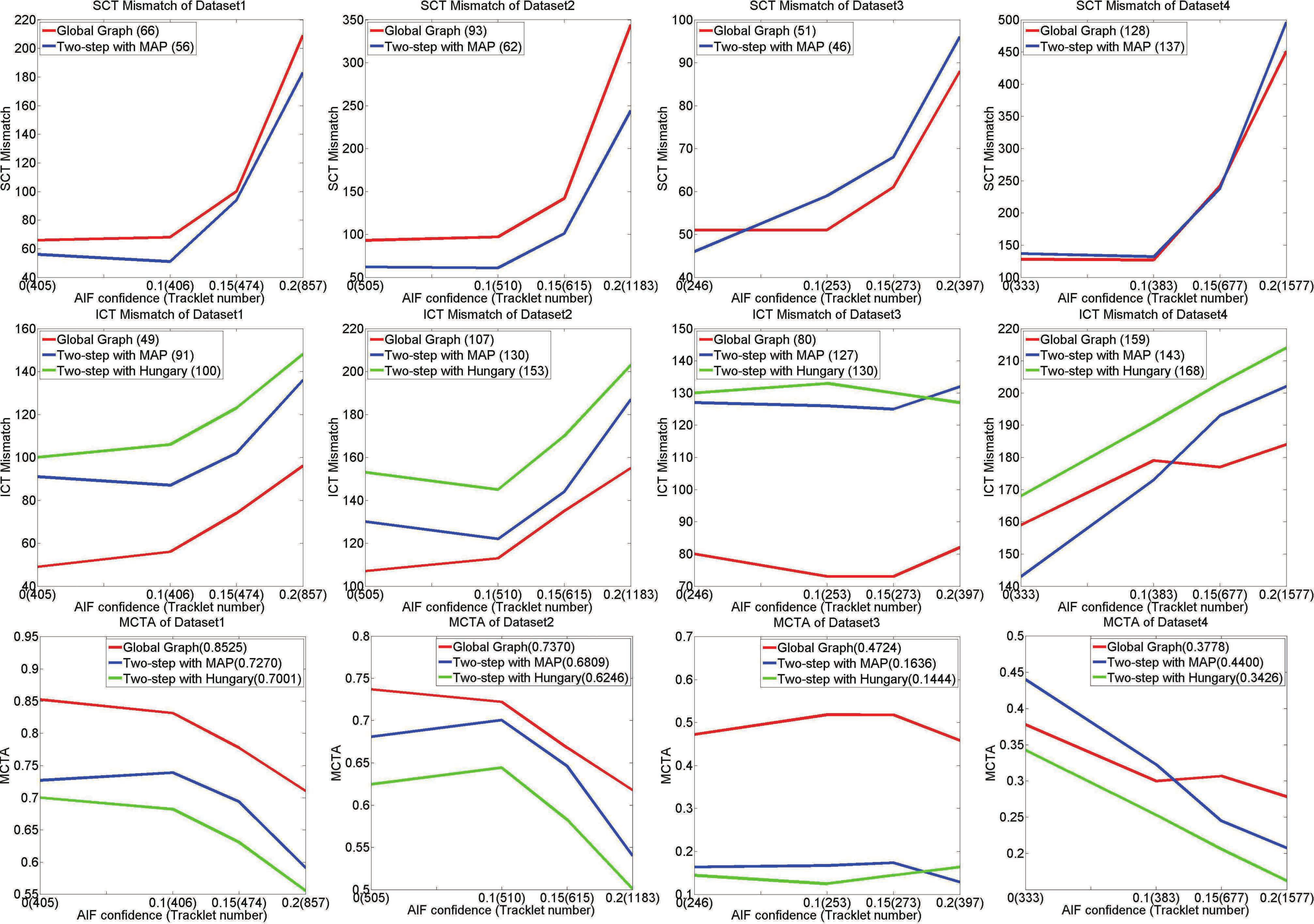}}
%
\caption{\textbf{Performance evaluation of the proposed approach under different parameter settings.} The x-coordinate for all the figures is the confidence threshold $\theta$ of the AIF tracker, and the number in bracket is the corresponding number of tracklets. With the increase of $\theta$, the tracklet number grows and more tracklet fragments are produced. The y-coordinates in three rows are the SCT mismatch number, the ICT mismatch number and the MCTA score respectively. The performance score under $\theta=0$ is shown in the legend. The method of global graph is the proposed approach. The two-step with MAP is Zhang's work \cite{MAP} which uses MAP to achieve the SCT process. The two-step with MAP and Hungary in the last two row stand for the approaches that solve the ICT problem with MAP and Hungary algorithm \cite{HungaryAlg}.}
\label{fig:Exp2}
\end{figure*}

\subsection{Evaluation Criteria}
\label{ssec:criteria}

As we know, both SCT and ICT have their own evaluation criteria. Most SCT
trackers usually use the multi-object tracking accuracy (MOTA) and
ID switch~\cite{MOTA} as their evaluation criteria, while some SCT papers
prefer other terms~\cite{Kuo/cvpr2011,Li/cvpr2009,Kuo/eccv2010}. In ICT, the ID switch is also
a necessary term.

There are two criteria mentioned in Section~\ref{sec:introduction} which are
important to a multi-camera multi-object tracking system. The SCT module and
the ICT module correspond to the two criteria respectively. As these two criteria are
equally crucial for multi-camera object tracking performance, they should be
considered equally important in the final performance measurement.

Nevertheless, in today's multi-camera object tracking, there is rarely a
widely accepted performance measurement that takes these two criteria into
account. The common criterion researchers used for multi-camera object tracking is an
extension of MOTA. It
adds the ID switches in SCT and in ICT together, which ignores the different incidence densities
of the ID switches in SCT and ICT. In most video scenes, \ie Table~\ref{table:Gth}, the ground truthes used for
frame matching in SCT are much more than those in ICT. It leads to trackers caring
more about the trajectories in single camera rather than the
inter-camera matching. In this paper, we treat them separately and
provide a new evaluation criterion to measure the performance of
multi-camera object tracking. Our criterion takes both of SCT and ICT criteria into account and uniform them into one
evaluation metric. The metric is called multi-camera object tracking
accuracy (MCTA):

\begin{equation}
\begin{array}{ll}
MCTA=Detection*Tracking^{SCT}*Tracking^{ICT} \\
\phantom{MCTA}=(\frac{2*Precision*Recall}{Precision+Recall})(1\!-\!\frac{\sum_tmme^s_t}{\sum_ttp^s_t})(1\!-\!\frac{\sum_tmme^c_t}{\sum_ttp^c_t}).
\end{array}
\label{eq:MCTA}
\end{equation}

It's also modified based on MOTA~\cite{MOTA} and can be applied on
multi-camera object tracking. It avoids the disadvantage of MOTA
that can be negative due to the false positives. The MCTA ranges
for 0 to 1. The metric contains three parts: detection ability, SCT
ability and ICT ability, which are corresponding to the three brackets in Eq.
\ref{eq:MCTA}. The $Precision$ and $Recall$ are integrated by
F1-score to measure the detection power and the occlusion handling
ability. In this paper, the experiments focus on testing the SCT and
the ICT abilities of the proposed approach, so for the first two experiments, we
use the ground truthes of object detections as the inputs instead of
running a real detector, which leads to $Precision=1$ and $Recall=1$. In the last experiment, a DPM \cite{Felzenszwalb/pami2010} detector is used to get the detection results.

\begin{equation}
\begin{array}{lll}
Detection=\frac{2*Precision*Recall}{Precision+Recall}, \\
Precision=1-\frac{\sum_tfp_t}{\sum_tr_t}, \\
Recall=1-\frac{\sum_tm_t}{\sum_tg_t},
\end{array}
\end{equation}
where $fp_t$, $r_t$, $m_t$ and $g_t$ are the number of false
positives, hypothesises, misses and ground truthes respectively in time $t$.

\begin{equation}
\begin{array}{ll}
Tracking^{SCT}=1-\frac{\sum_tmme^s_t}{\sum_ttp^s_t}, \\
Tracking^{ICT}=1-\frac{\sum_tmme^c_t}{\sum_ttp^c_t}.
\end{array}
\label{eq:TrackingTerm}
\end{equation}

For SCT and ICT ability parts, we measure the abilities via the
number of mismatches (ID-switches). We split the number of
mismatches $mme_t$ in MOTA~\cite{MOTA} into $mme^s_t$ and $mme^c_t$.
$mme^s_t$ represents the number of mismatches happened in a single
camera and $mme^c_t$ is for those inter-camera mismatches. The
$tp^s_t$ and $tp^c_t$ are the matching numbers of frames
in ground truthes. $tp^s_t$ contains the
matchings, the two frames of which are from the
same camera, and $tp^c_t$ means the number of those inter-camera
matchings. It is worth noting that both $tp^s_t$ and $tp^c_t$ are among the truth positive detection results. For a new target, it's counted as an inter-camera
ground truth by default in our criterion.

\begin{table}\scriptsize
\caption{Empirical comparison of the proposed approach on four multi-camera tracking datasets. The bold indicates the best performance.}
\renewcommand{\arraystretch}{1.5}
\begin{center}
\resizebox{\linewidth}{!}{
\begin{tabular} {|c|c|c|c|c|c|}
  \hline
    \multicolumn{2}{|c|}{} & NonA & EqlA & M & EqlA+M \\
  \hline
    \multirow{3}{*}{Dataset1} & $mme^s$ & 71 & 76 & {\bf 53} & 66\\
  \cline{2-6}
    & $mme^c$ & 123 & 88 & 101 & {\bf 49}\\
  \cline{2-6}
    & $MCTA$ & 0.6311 & 0.7357 & 0.6971 & {\bf 0.8525}\\
  \hline
    \multirow{3}{*}{Dataset2} & $mme^s$ & 83 & 109 & {\bf 67} & 93 \\
  \cline{2-6}
    & $mme^c$ & 201 & 164 & 126 & {\bf 107}\\
  \cline{2-6}
    & $MCTA$ & 0.5069 & 0.5973 & 0.6907 & {\bf 0.7370}\\
  \hline
    \multirow{3}{*}{Dataset3} & $mme^s$ & 59 & 71 & 74 & {\bf 51}\\
  \cline{2-6}
    & $mme^c$ & 132 & 116 & 95 & {\bf 80}\\
  \cline{2-6}
    & $MCTA$ & 0.1312 & 0.2359 & 0.3735 & {\bf 0.4724}\\
  \hline
    \multirow{3}{*}{Dataset4} & $mme^s$ & 125 & 137 & {\bf 123} & 128\\
  \cline{2-6}
    & $mme^c$ & 187 & 169 & 188 & {\bf 159}\\
  \cline{2-6}
    & $MCTA$ & 0.2687 & 0.3388 & 0.2649 & {\bf 0.3778}\\
  \hline
    \multicolumn{2}{|c|}{$Average MCTA$} & 0.3845 & 0.4769 & 0.5066 & {\bf 0.6099}\\
  \hline
\end{tabular}}
\end{center}
\label{table:E1}
\end{table}

\subsection{Global Graph Model vs Two-Step Framework}
\label{ssec:Exp4GlobalMAP}

The advantage of the proposed method is to improve the ICT
performance under an unperfect SCT result. So in this section, the
proposed global graph model is compared with the traditional
two-step framework, \ie a SCT approach plus an ICT approach. We use the
same MAP model to solve the data association in both SCT and ICT
steps in the two-step framework and aim to remove the
interference of different data association methods. Adopting the MAP
model in SCT is presented in Zhang~\etal~\cite{MAP}. However using
MAP model in ICT is not a suitable solution when the tracking
results in single camera are perfect and unchangeable. But as we
said in Section.~\ref{ssec:ICT}, when the SCT results are not ideal,
the data association in ICT should be more like a global
optimization problem rather than a K-partite graph matching problem, which can be solved by the MAP model.
That's another reason why we use the MAP model to achieve the data
association in ICT in the traditional two-step framework.
As a complement, we also utilize Hungary
algorithm~\cite{HungaryAlg} to achieve the ICT step, which
is a classical data association method for ICT.
The feature representation in this experiment is the PMCSHR
appearance and motion features for all baselines due to the fairness reason.

In this experiment, the waiting time threshold $\eta$ and the
minimum value $\varepsilon$ of the MUG are set to 60*25*1 and 0.4
respectively, the weights of two features $k_1$ and $k_2$ are both
1. To prove the ability of the proposed approach handling unperfect
tracklets in SCT, the experiment changes the threshold $\theta$ of
the confidence of the AIF tracker to produce more fragments
artificially. The threshold $\theta$ ranges from 0 to 0.2 and the
corresponding numbers of tracklets are listed beside the threshold in Fig. \ref{fig:Exp2}.

The total single-camera matching number $tp^s$ and inter-camera
matching number $tp^c$ of ground truthes for each sub-dataset are listed in Table
\ref{table:Gth}. From the first two rows in Fig.~\ref{fig:Exp2}, we can see that with the
increase of the fragmented tracklet number, both the single camera mismatch number
$mme^s$ and the inter-camera mismatch number $mme^c$ grow significantly
in the proposed global graph and the two-step framework. In the first row, the single
camera mismatch number $mme^s$ in the proposed global graph is always
larger than that in the two-step framework~\cite{MAP}, because the two-step framework
offers an optimization in each camera which makes it have a better
local result. In dataset 3 and dataset 4, the $mme^s$ in the
proposed global graph becomes lower than that in the two-step
framework~\cite{MAP}. The reason is that these two datasets are under
a simulation condition which have many frequent ``walking around''
behaviors. In this case, the inter-camera information may
provide more useful feedbacks for each specific camera and can
partly improve the SCT performance. For the inter-camera mismatch
number $mme^c$ in the middle row, the number in the proposed global graph is much
lower than that in both MAP and Hungary graph~\cite{HungaryAlg} in the two-step
framework, it indicates the effectiveness of our global graph model
to improve the ICT performance. In dataset 4, it can be seen that
the $mme_c$ in the proposed graph is not smaller than that in the two-step framework at first time. However, with the increase of fragmented tracklets, the
$mme_c$ in the proposed graph increases much more slowly and
finally becomes smaller than that in the two-step
framework. What's more, as the ICT step in two-step framework, the data association method
based on the global MAP is always better than that with Hungary algorithm~\cite{HungaryAlg}. It can partly prove the assumption that the data association in ICT is more suitable to be treat as a global optimization problem rather than a K-partite graph matching problem because of non-ideal SCT results.
In the last row, the MCTA of the global MAP always keep the highest score,
which implies that the proposed global graph model offers a better
performance compared with the traditional two-step framework.

\begin{table}
\renewcommand{\arraystretch}{1.5}
\caption{Performance comparison using the ground truthes of single camera object tracking as input.}
\begin{center}
\resizebox{\linewidth}{!}{
\begin{tabular} {|c|c|c|c|c|c|}
  \hline
    \multicolumn{2}{|c|}{} & Ours & USC-Vision & Hfutdspmct  & CRIPAC-MCT  \\
    \multicolumn{2}{|c|}{} & & \cite{USCVision1}+\cite{USCVision2} & \cite{MCTchallenge} & \cite{Chen/icip2014} \\
  \hline
    \multirow{2}{*}{Dataset1} & $mme^c$ & 55 & 27 & 86 & 113\\
  \cline{2-6}
    & $MCTA$ & 0.8353 & 0.9152 & 0.7425 & 0.6617 \\
  \hline
    \multirow{2}{*}{Dataset2} & $mme^c$ & 121 & 34 & 141 & 167 \\
  \cline{2-6}
    & $MCTA$ & 0.7034 & 0.9132 & 0.6544 & 0.5907 \\
  \hline
    \multirow{2}{*}{Dataset3} & $mme^c$ & 39 & 70 & 40 & 44 \\
  \cline{2-6}
    & $MCTA$ & 0.7417 & 0.5163 & 0.7368 & 0.7105 \\
  \hline
    \multirow{2}{*}{Dataset4} & $mme^c$ & 157 & 72 & 155 & 110 \\
  \cline{2-6}
    & $MCTA$ & 0.3845 & 0.7052 & 0.3945 & 0.5703 \\
  \hline
    \multicolumn{2}{|c|}{$Average MCTA$} & 0.6662 & 0.7625 & 0.6321 & 0.6333 \\
  \hline
\end{tabular}}
\end{center}
\label{table:E3}
\end{table}

\subsection{Equalized vs Non-equalized Graph Model}
\label{ssec:Exp4similarity}

This experiment is conducted to prove the effectiveness of the
similarity equalization process. All the trackers are under our
global graph model. We compare the equalized appearance similarity metric
with the non-equalized one and then combined with our equalized
motion metric. Particularly, in this experiment, the confidence threshold $\theta$ of the AIF tracker is fixed and set to 0.

The results are shown on Table \ref{table:E1}.
NonA and EqlA are the results with the non-equalized and the equalized appearance features. M is corresponding to the results with the equalized motion feature
only and EqlA+M is the one that combines the equalized appearance feature and
the motion feature together.
It can be found that the result with the non-equalized appearance similarity has a
lower mismatch number $mme^s$ in the single camera compared with the equalized one. It means that when we conduct
equalization, the single camera performance drops down due to the
change of the distribution of the single camera similarity, and that is
unavoidable but acceptable. In the inter-camera tracking, it is clear that
the equalized appearance similarity tracker gives a great help to
reduce the number $mme^c$ of mismatches across cameras. When the
equalized motion information is added in, the $mme^c$ further
decreases. The MCTA is the final comprehensive score which takes both SCT and ICT performances into
account. The larger the score is, the better performance the tracker
has. As seen in Table \ref{table:E1}, the equalized
appearance similarity result combined with the equalized motion
information has a highest score. It indicates that the increased
single camera mismatch number $mme^s$ in our method is acceptable in
order to reduce the inter-camera mismatch number $mme^c$ and get a
higher score in the whole MCT performance. Further more, when we use
the motion feature alone for the multi-camera object tracking, the
performance is comparable and sometimes better than the appearance feature, which partly proves the effectiveness of our equalized motion similarity metric.

\begin{table}
\renewcommand{\arraystretch}{1.5}
\caption{Performance comparison using the ground truthes of object detection as input.}
\begin{center}
\resizebox{\linewidth}{!}{
\begin{tabular} {|c|c|c|c|c|c|}
  \hline
    \multicolumn{2}{|c|}{} & Ours & USC-Vision & Hfutdspmct  & CRIPAC-MCT  \\
    \multicolumn{2}{|c|}{} & & \cite{USCVision1}+\cite{USCVision2} & \cite{MCTchallenge} & \cite{Chen/icip2014} \\
  \hline
    \multirow{3}{*}{Dataset1} & $mme^s$ & 66 & 63 & 77 & 135 \\
  \cline{2-6}
    & $mme^c$ & 49 & 35 & 84 & 103\\
  \cline{2-6}
    & $MCTA$ & 0.8525 & 0.8831 & 0.7477 & 0.6903 \\
  \hline
    \multirow{3}{*}{Dataset2} & $mme^s$ & 93 & 61 & 109 & 230 \\
  \cline{2-6}
    & $mme^c$ & 107 & 59 & 140 & 153 \\
  \cline{2-6}
    & $MCTA$ & 0.7370 & 0.8397 & 0.6561 & 0.6234 \\
  \hline
    \multirow{3}{*}{Dataset3} & $mme^s$ & 51 & 93 & 105 & 147 \\
  \cline{2-6}
    & $mme^c$ & 80 & 111 & 121 & 139 \\
  \cline{2-6}
    & $MCTA$ & 0.4724 & 0.2427 & 0.2028 & 0.0848 \\
  \hline
    \multirow{3}{*}{Dataset4} & $mme^s$ & 128 & 70 & 97 & 140\\
  \cline{2-6}
    & $mme^c$ & 159 & 141 & 188 & 209 \\
  \cline{2-6}
    & $MCTA$ & 0.3778 & 0.4357 & 0.2650 & 0.1830 \\
  \hline
    \multicolumn{2}{|c|}{$Average MCTA$} & 0.6099 & 0.6003 & 0.4679 & 0.3954 \\
  \hline
\end{tabular}}
\end{center}
\label{table:E4}
\end{table}

\subsection{Equalized Global Graph Model vs State of The Arts}
\label{ssec:Exp4Others}

\begin{figure*}[t]
\begin{minipage}[b]{0.45\linewidth}
  \centering
  \includegraphics[width=1\linewidth]{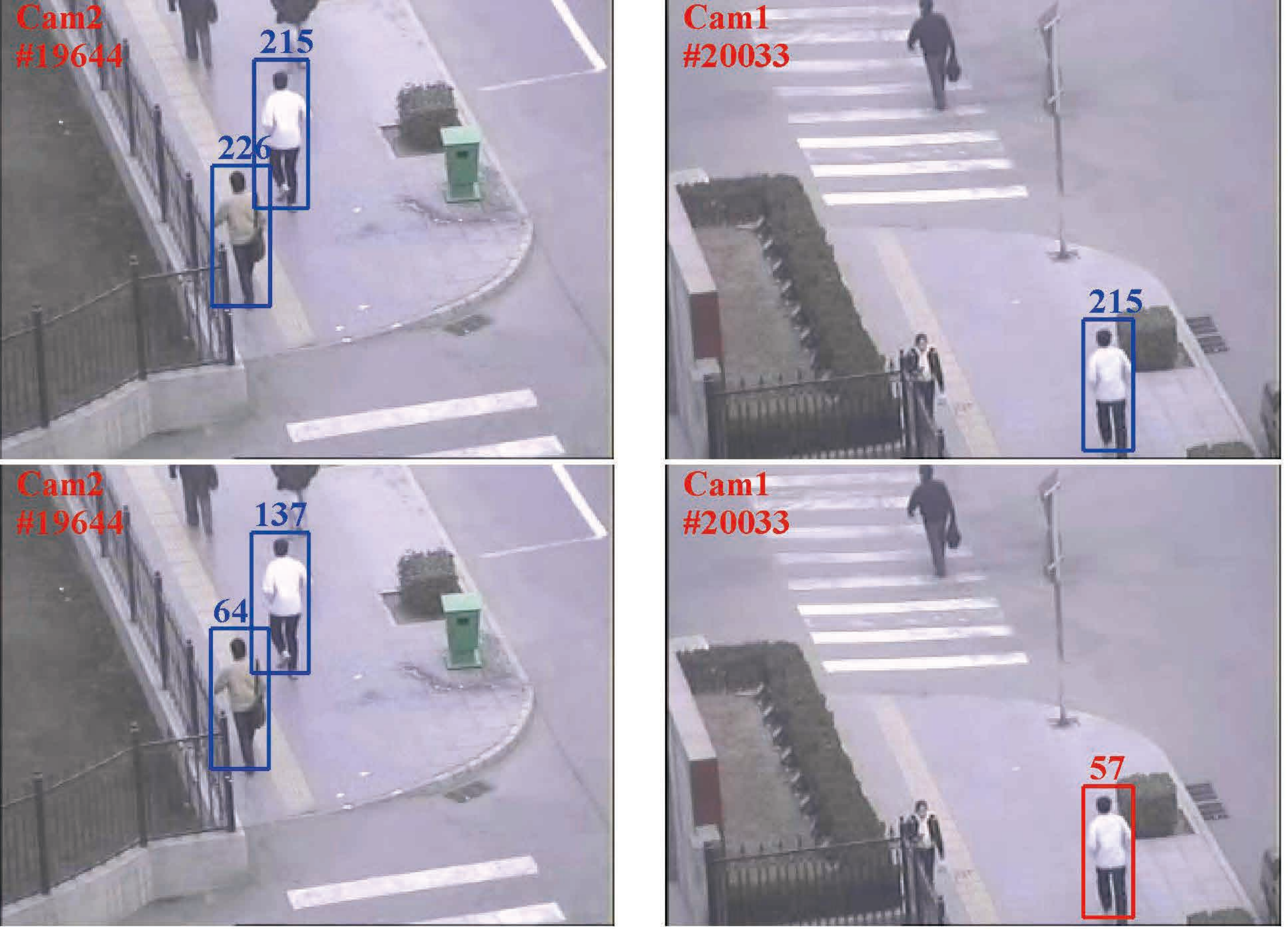}
  (a)Dataset 1: Outdoor scene
\end{minipage}
\hfill
\begin{minipage}[b]{0.45\linewidth}
  \centering
  \includegraphics[width=1\linewidth]{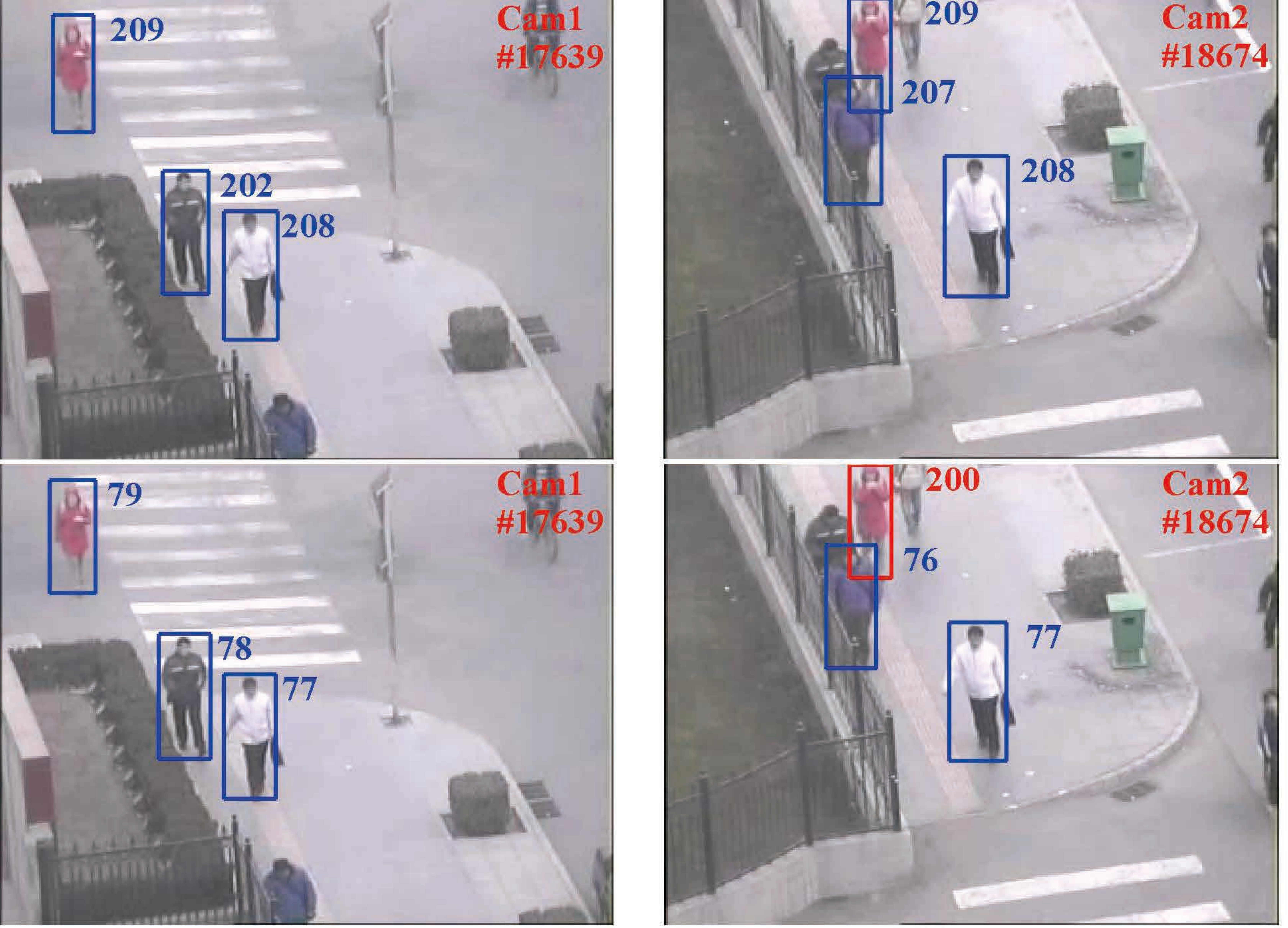}
  (b)Dataset 2: Outdoor scene
\end{minipage}
\hfill
\begin{minipage}[b]{0.45\linewidth}
  \centering
  \includegraphics[width=1\linewidth]{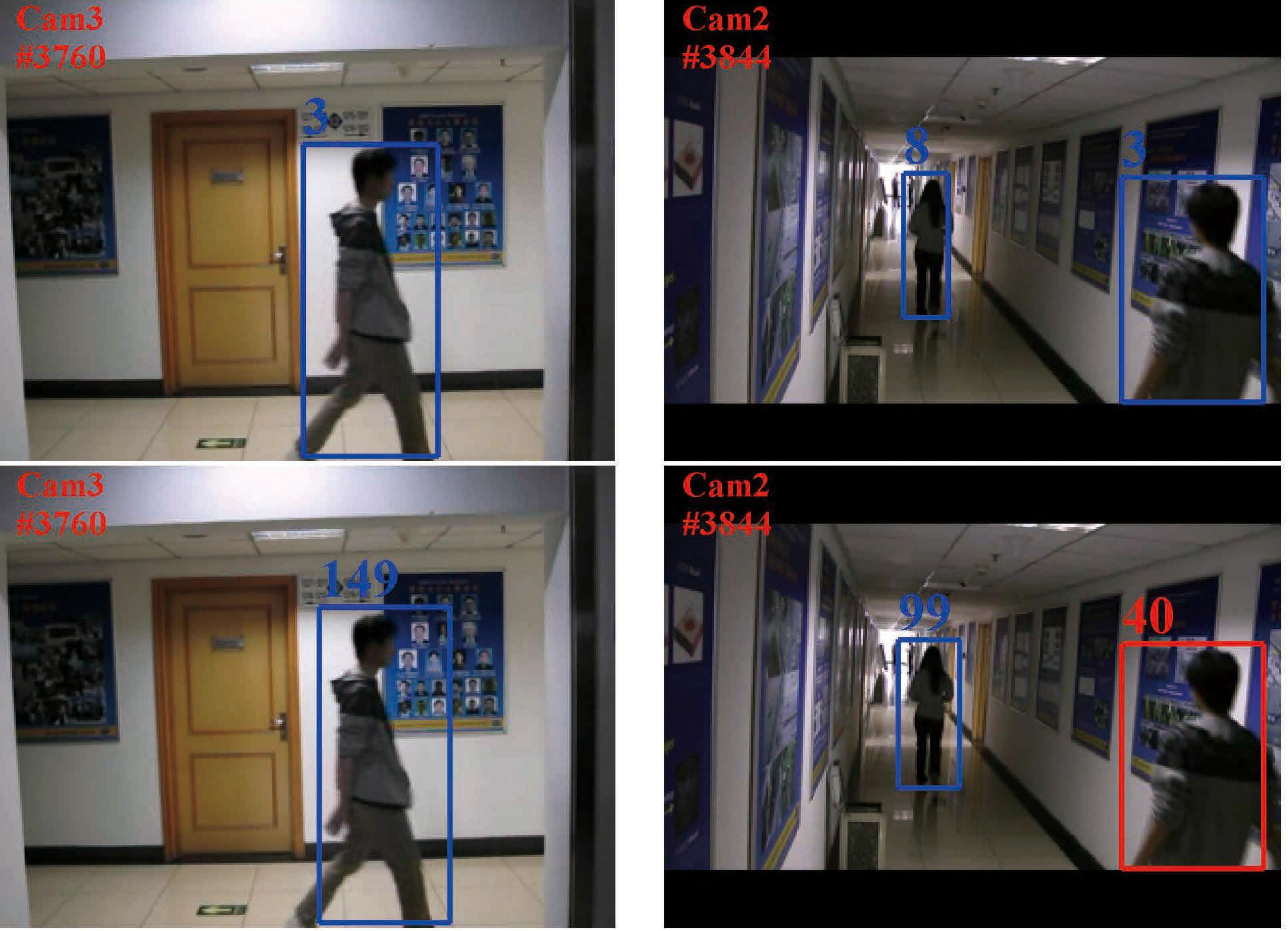}
  (c)Dataset 3: Indoor scene
\end{minipage}
\hfill
\begin{minipage}[b]{0.45\linewidth}
  \centering
  \includegraphics[width=1\linewidth]{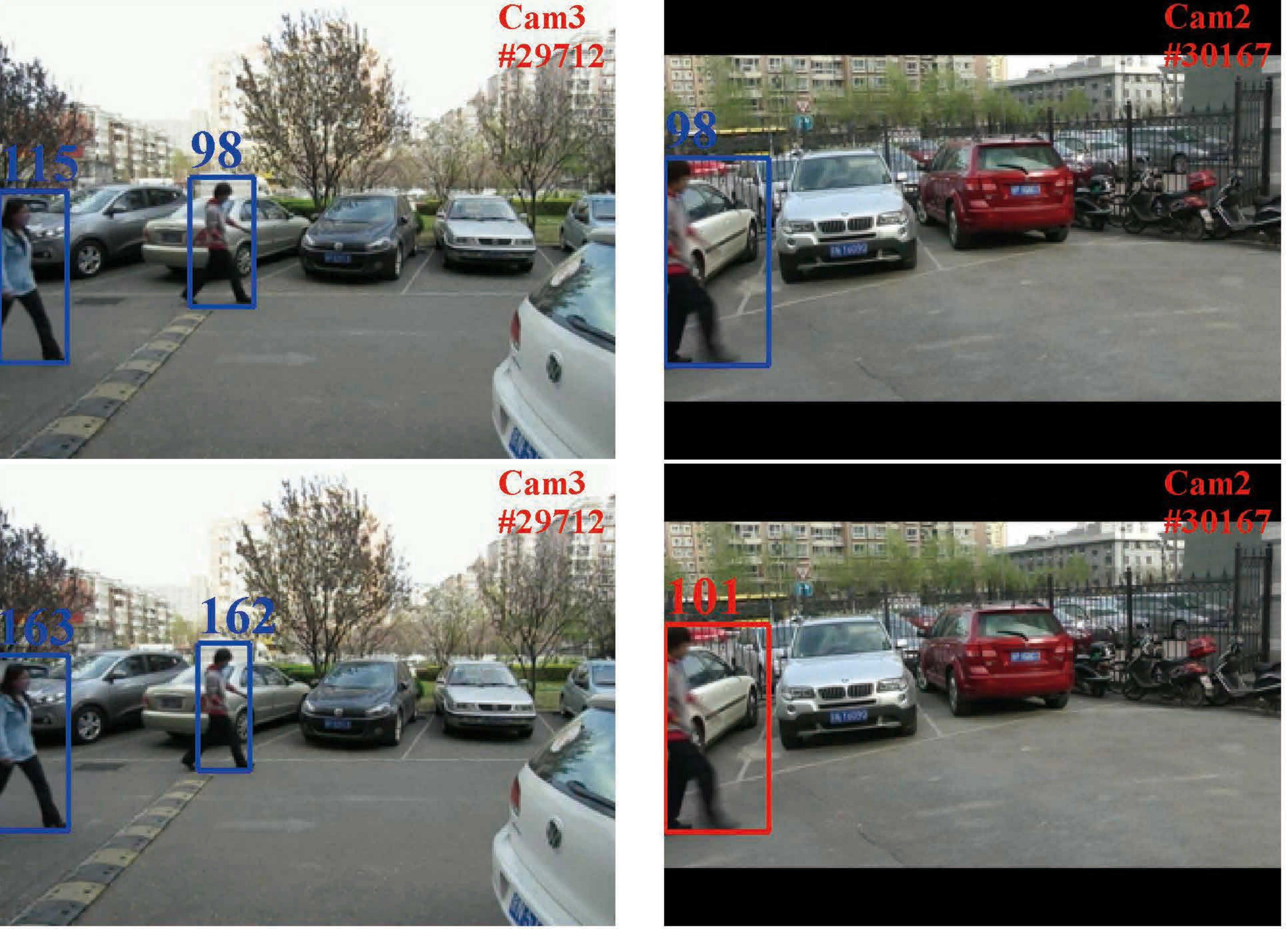}
  (d)Dataset 4: outdoor scene
\end{minipage}
\caption{\textbf{Samples of mismatches in the inter-camera tracking.} Four inter-camera tracking examples from Dataset 1-4 are shown in (a)-(d). The first row is the results of the proposed method, and the second row is from USC-Vision (\cite{USCVision1,USCVision2}). The red rect means the mismatch happened across cameras.}
\label{fig:Mismatch}
\end{figure*}

In this section, we compare our equalized global MAP graph model
with other multi-camera object tracking methods. As a comparison, the methods must contain the abilities to handle both the SCT
and the ICT steps. We compare the proposed graph with current
two-step multi-camera object tracking methods. The methods are from the Multi-Camera Object
Tracking (MCT) Challenge~\cite{MCTchallenge}. USC-Vision
(~\cite{USCVision1,USCVision2}) is the winner in the challenge
which is considered as the state-of-the-art two-step multi-camera object tracking approach. We first conduct the
comparison under the condition that the ground truthes of single
camera object tracking are available, the results are shown in Table \ref{table:E3}. It reflects the ICT power of each method when the
single camera object tracking results are perfect. From the average
MCTA score we can see that USC-Vision
(\cite{USCVision1,USCVision2}) is much better than our
proposed method. This proves the advantage of USC-Vision's ICT
method. In Table \ref{table:E4}, only the ground truthes of object
detections are available, the tracker should achieve the single camera
object tracking by themselves.
On this occasion, their results of the single camera object tracking can't be as perfect as the ground truthes, and
their inter-camera object tracking algorithms have to bear these fragments and false positives.
From Table \ref{table:E4}, although the SCT
performance $mme^s$ of USC-Vision
(\cite{USCVision1,USCVision2}) is better than ours, it is clear that the number of its ICT mismatches increases much
more shapely than our method's, which indicates that its powerful
ICT method loses its advantage under the unperfect SCT results. Results are shown in Fig. \ref{fig:Mismatch}. As
the final evaluation, our equalized global graph model has the
highest average MCTA score, which further proves the advantage of our
proposed model on improving the ICT performance under an unperfect SCT
result. At last, as perfect detection can never be achieved in reality, we do another experiment without the detection ground truthes. We uses the DPM detector \cite{Felzenszwalb/pami2010} to get the detection results. In Table~\ref{table:E5} the $Tracking^{SCT}$ and $Tracking^{ICT}$
corresponding to Eq.~\ref{eq:TrackingTerm} are listed instead of $mme$ because the different detection results would cause different $tp$s.
From the results in Table~\ref{table:E5}, it shows that our result is not the best but can be comparable with the state of the arts.
Under a real detector, there would be much missing and false positive detections. The ability of a multi-camera tracker to handle these missing or false positive detections mainly comes from its SCT part. USC-Vision uses a hierarchical association to build its tracklets, in which the detections are selected discreetly and some missing detections can be partly complemented. In our method, a real-time single object tracker~\cite{AIF} is adopted to get the tracklets, which can partly handle missing detections. But for the false detections, once the tracker drifts to a false detection, it would cause the whole tracklet unreliable. Due to the benefits of the hierarchical association in the SCT step, USC-Vision has a more reliable set of tracklets than those we have for the next ICT step. Even with the help of the proposed equalized global graph, our final result is still a little lower than USC-Vision's. This can't deny the effectiveness of our equalized global graph model, but prove the advantage of USC-Vision's SCT method to handle misses and false positives. However, for practical usages in real environment, the detection-level association is much slower than a real-time single camera tracker. That's why we use the AIF tracker to get tracklets in our method instead of using USC-Vision's detection-based hierarchical association. Some other single object trackers, such as TLD~\cite{TLD}, may handle the false-detection drifts by their online learning mechanisms. But it costs too much time and memories on learning the online models, which is hard to be applied on forming our raw tracklets. As a result, a real-time single camera tracker that can deal with the false detections is a promising further work for multi-camera object tracking.

\begin{table}
\renewcommand{\arraystretch}{1.5}
\caption{Performance comparison without the ground truthes of object detection. The final MCTA is shown as bold for clarity.}
\begin{center}
\resizebox{\linewidth}{!}{
\begin{tabular} {|c|c|c|c|c|c|}
  \hline
    \multicolumn{2}{|c|}{} & Ours & USC-Vision & Hfutdspmct  & CRIPAC-MCT  \\
    \multicolumn{2}{|c|}{} & & \cite{USCVision1}+\cite{USCVision2} & \cite{MCTchallenge} & \cite{Chen/icip2014} \\
  \hline
    \multirow{5}{*}{Dataset1} & $precision$ & 0.7967 & 0.6916 & 0.7113 & 0.1488 \\
  \cline{2-6}
    & $recall$ & 0.5929 & 0.6061 & 0.3465 & 0.2154\\
  \cline{2-6}
    & $Tracking^{SCT}$ & 0.9744 & 0.9981 & 0.9229 & 0.9955\\
  \cline{2-6}
    & $Tracking^{ICT}$ & 0.6220 & 0.9288 & 0.6534 & 0.7111\\
  \cline{2-6}
    & $MCTA$ & {\bf 0.4120} & {\bf 0.5989} & {\bf 0.2810} & {\bf 0.1246} \\
  \hline
    \multirow{5}{*}{Dataset2} & $precision$ & 0.7977 & 0.6948 & 0.7461 & 0.1431 \\
  \cline{2-6}
    & $recall$ & 0.6332 & 0.7843 & 0.3669 & 0.1933\\
  \cline{2-6}
    & $Tracking^{SCT}$ & 0.9779 & 0.9986 & 0.9347 & 0.9945\\
  \cline{2-6}
    & $Tracking^{ICT}$ & 0.6942 & 0.8507 & 0.6122 & 0.7510 \\
  \cline{2-6}
    & $MCTA$ & {\bf 0.4793} & {\bf 0.6260} & {\bf 0.2815} & {\bf 0.1075} \\
  \hline
    \multirow{5}{*}{Dataset3} & $precision$ & 0.8207 & 0.4750 & 0.3342 & 0.0853 \\
  \cline{2-6}
    & $recall$ & 0.5345 & 0.6615 & 0.0986 & 0.1206\\
  \cline{2-6}
    & $Tracking^{SCT}$ & 0.9749 & 0.9904 & 0.9682 & 0.9715\\
  \cline{2-6}
    & $Tracking^{ICT}$ & 0.2953 & 0.1014 & 0.2432 & 0.1143 \\
  \cline{2-6}
    & $MCTA$ & {\bf 0.1864} & {\bf 0.0555} & {\bf 0.0359} & {\bf 0.0111} \\
  \hline
    \multirow{5}{*}{Dataset4} & $precision$ & 0.8355 & 0.5216 & 0.7720 & 0.0606\\
  \cline{2-6}
    & $recall$ & 0.6193 & 0.79375 & 0.1210 & 0.0944\\
  \cline{2-6}
    & $Tracking^{SCT}$ & 0.9275 & 0.9948 & 0.9865 & 0.9762\\
  \cline{2-6}
    & $Tracking^{ICT}$ & 0.4308 & 0.5437 & 0.2944 & 0.2950 \\
  \cline{2-6}
    & $MCTA$ & {\bf 0.2842} & {\bf 0.3404} & {\bf 0.0608} & {\bf 0.0213} \\
  \hline
    \multicolumn{2}{|c|}{$Average MCTA$} & {\bf 0.3405} & {\bf 0.4052} & {\bf 0.1648} & {\bf 0.0661} \\
  \hline
\end{tabular}}
\end{center}
\label{table:E5}
\end{table}

\section{Conclusion}
In order to address the problem of multi-camera non-overlapping visual object tracking, we develop a
joint approach that optimising the single camera object tracking and the inter-camera object tracking in one graph.
This joint approach overcomes the disadvantages in the traditional two-step tracking approaches.
In addition, the similarity metrics of both appearance and motion
features in the proposed global graph are equalized.
The equalization further reduces the number of mismatch errors in
inter-camera object tracking.
The results show its effectiveness for multi-camera object tracking,
especially when the SCT performance is not perfect.
Our approach focuses on the graph modeling instead of
the feature representation learning. Any existing re-identification
feature representation method can be incorporated
into our framework.

\ifCLASSOPTIONcaptionsoff
  \newpage
\fi

\bibliographystyle{IEEEtran}
\bibliography{refs}

\end{document}